\documentclass[journal]{IEEEtran}
\usepackage{amsmath,amsfonts}
\usepackage{algorithm,algpseudocode}
\usepackage{marvosym}
\usepackage{makecell}
\usepackage[normalem]{ulem}
\usepackage{array}
\usepackage[caption=false,font=normalsize,labelfont=sf,textfont=sf]{subfig}
\usepackage{textcomp}
\usepackage{stfloats}
\usepackage{url}
\usepackage{verbatim}
\usepackage{graphicx}
\usepackage{cite}
\usepackage{amssymb}
\usepackage{booktabs}
\usepackage{multirow}
\usepackage[table]{xcolor}
\usepackage{colortbl}
\usepackage[capitalize]{cleveref}
\Crefname{section}{Section}{Sections}
\Crefname{table}{Table}{Tables}
\Crefname{figure}{Figure}{Figures}
\definecolor{my_color}{HTML}{e8eef1}

\ifCLASSINFOpdf
\else
\fi
\begin{document}

\title{DSwinIR: Rethinking Window-based Attention for Image Restoration}

\author{Gang Wu,~\IEEEmembership{Student Member,~IEEE,}
        ~Junjun Jiang,~\IEEEmembership{Senior Member,~IEEE,}
        ~Kui~Jiang,~\IEEEmembership{Member,~IEEE,}
        ~Xianming~Liu,~\IEEEmembership{Member,~IEEE,}
        ~and~Liqiang Nie,~\IEEEmembership{Senior Member,~IEEE}
        
 \thanks{The research was supported by the National Natural Science Foundation of China (U23B2009, 62471158) and the Natural Science Foundation of Heilongjiang Province of China for Excellent Youth Project (YQ2024F006).} 
\IEEEcompsocitemizethanks{

\IEEEcompsocthanksitem  G. Wu, J. Jiang, K. Jiang, and X. Liu are with the School of Computer Science and Technology, Harbin Institute of Technology, Harbin 150001, China. E-mail: \{gwu@hit.edu.cn, jiangjunjun@hit.edu.cn, jiangkui@hit.edu.cn, csxm@hit.edu.cn\}. Corresponding author: Junjun Jiang.
\IEEEcompsocthanksitem L. Nie is with the School of Computer Science and Technology, Harbin Institute of Technology (Shenzhen), Shenzhen 518055, China. E-mail: \{nieliqiang@gmail.com\}.

}}
        


\maketitle

\IEEEpeerreviewmaketitle

\begin{abstract}
Image restoration has witnessed significant advancements with the development of deep learning models.
Transformer-based models, particularly those using window-based self-attention, have become a dominant force. However, their performance is constrained by the rigid, non-overlapping window partitioning scheme, which leads to \textit{insufficient feature interaction across windows and limited receptive fields}. This highlights the need for more adaptive and flexible attention mechanisms.
In this paper, we propose the Deformable Sliding Window Transformer for Image Restoration (DSwinIR), a new attention mechanism: the {Deformable Sliding Window (DSwin) Attention}. {This mechanism introduces a token-centric and content-aware paradigm that moves beyond the grid and fixed window partition.} It comprises two complementary components. First, it replaces the rigid partitioning with a \textit{token-centric sliding window} paradigm, {making it effective at eliminating boundary artifacts}. Second, it incorporates a \textit{content-aware deformable sampling} strategy, which allows the attention mechanism to learn data-dependent offsets and actively shape its receptive field to focus on the most informative image regions. Extensive experiments show that DSwinIR achieves strong results, including state‑of‑the‑art performance on several evaluated benchmarks. For instance, in all-in-one image restoration, our DSwinIR surpasses the most recent backbone GridFormer by \textbf{0.53 dB} on the three-task benchmark and \textbf{0.87 dB} on the five-task benchmark.
The code and pre-trained models are available at \textit{https://github.com/Aitical/DSwinIR}.
\end{abstract}


\begin{figure}[h] 
\centering \includegraphics[width=0.475\textwidth]{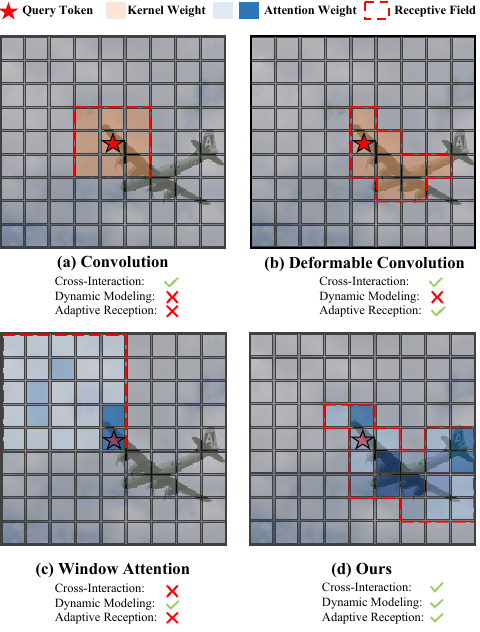} 
\caption{Comparative analysis of feature extraction mechanisms with an anchor token (marked by {\large{\textcolor{red}{$\star$}}}) as the reference point. (a) Vanilla convolution applies a fixed sampling pattern, leveraging neighborhood features. (b) Deformable convolution introduces adaptive sampling locations based on content, enabling more effective feature integration from relevant regions. (c) Window attention suffers from boundary constraints where anchor tokens near window edges (especially corners) have limited receptive fields. (d) The proposed Deformable Sliding Window (DSwin) attention extends window attention with a token-centric paradigm and the content-aware sampling, bringing robust feature aggregation for anchor tokens.}

\vspace{-4mm}
\label{fig_intro_dswin} 
\end{figure}

\section{Introduction}
\IEEEPARstart{I}{mage} restoration aims to recover high‑quality images from inputs degraded by noise, rain, haze, blur, low light, and other factors~\cite{survey_denoise_Michael2023,survey_dehaze_gui2023,survey_derain_chen2023,Zhang22blurSurvey,survey_all_in_one_jiang2024}. Deep learning has led to substantial progress, initially with CNNs~\cite{DnCNN, zamir2021multi} and more recently, with Vision Transformers (ViTs)~\cite{ViT_Alexey_2021} that model long‑range dependencies.

\begin{figure*}[!t] 
\centering \includegraphics[width=\linewidth]{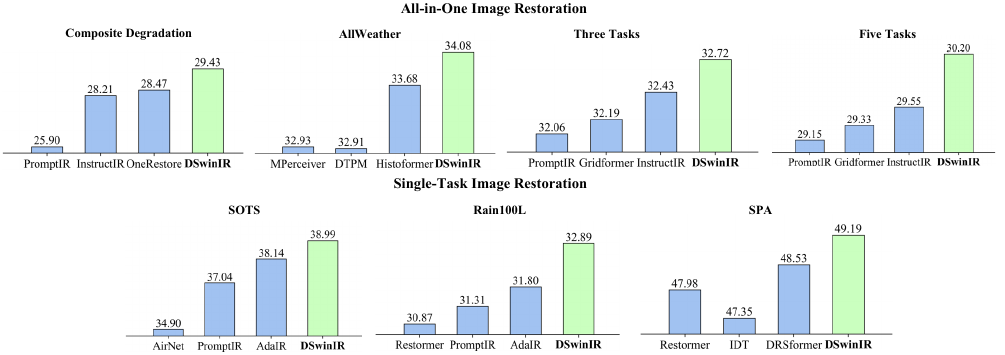} 
\caption{Quantitative comparison of the proposed DSwinIR against existing methods across diverse image restoration tasks, achieving consistent superior performance. All metrics are reported in PSNR (dB).} 
\label{fig_intro_compare} 
\end{figure*}

The practical use of vanilla Transformers for image restoration is impeded by the quadratic cost of global self-attention with respect to input resolution. Window-based self-attention \cite{SwinT_LiuL00W0LG21}, adopted in SwinIR \cite{SwinIR} and Uformer \cite{Uformer}, has therefore become the \textit{de facto} choice, striking a favorable trade-off between fidelity and efficiency by restricting computation to local, non-overlapping windows.
It introduces two salient limitations:
{\textit{1) Boundary Context Truncation}. Consider a token at a window corner: its receptive field is clipped precisely where informative context often lies across the boundary. As a result, its representation collapses toward its nearest in-window neighbors rather than the true cross-window structure it belongs to, producing suboptimal, biased features.}
\textit{2) Inflexible Receptive Fields:} The fixed, square-shaped attention pattern is applied uniformly across the image, failing to adapt to the diverse and anisotropic nature of image content and degradation patterns. 
{Subsequent efforts introduce hand-crafted interaction patterns, such as cross-shaped windows \cite{CAT_ChenZGzKY22}, axial stripes \cite{StripFormer_TsaiPLTL22}, and sparse token sampling \cite{DRSformer}. While effective, these essentially replace one fixed interaction pattern with another, leaving the core limitation unresolved.}
They fall short of addressing the core, underlying principle: the shape and scope of an attention field should be dynamically determined by the image content itself, not by a predefined, static template.

In this work, we move beyond such predefined, static patterns and advocate that an effective attention mechanism for image restoration is both \textit{token-centric} and \textit{content-aware}. Motivated by this view, we introduce Deformable Sliding Window (DSwin) Attention, a redesign of windowed attention (Figure~\ref{fig_intro_dswin}). First, to strengthen cross‑window interaction, we embrace a \textit{token-centric sliding window} paradigm: every token attends within its own centered neighborhood, creating overlaps that reduce boundary effects. Second, to tackle the inflexible receptive field, we introduce a \textit{content-aware deformable mechanism}: learned offsets adapt the sampling pattern to image structure (e.g., diagonal rain streaks or object contours). Building upon core components, we present the {Deformable Sliding Window Transformer for Image Restoration (DSwinIR)}. It integrates the DSwin attention module, offering a flexible and robust backbone for diverse image restoration tasks.

In recent years, the development of all-in-one image restoration models, handling multiple types of degradation within a unified framework, has gained increasing attention~\cite{survey_all_in_one_jiang2024}. We evaluate DSwinIR in the all‑in‑one setting. Strengthening the backbone architecture offers an underexplored alternative to prompt‑based approaches~\cite{PromptIR,cui2025adair}. Empirically, DSwinIR matches or exceeds several prompt‑based methods on our benchmarks.
We conduct extensive evaluations across diverse image restoration tasks, spanning both all-in-one multiple degradation scenarios and specialized single-task settings. As illustrated in Figure~\ref{fig_intro_compare}, DSwinIR delivers substantial improvements across diverse benchmarks.

The main contributions of this work are summarized as follows:
\begin{itemize}
    \item We propose a novel {Deformable Sliding Window (DSwin) Attention} mechanism. It adopts a token‑centric, content‑aware design that addresses boundary truncation and inflexible receptive fields in window‑based attention.
    
    \item We develop {DSwinIR}, a new and general backbone architecture for image restoration built upon the DSwin attention module. The architecture cohesively integrates multi-scale feature extraction in both the attention and feed-forward network modules, further enhancing its representation capacity.
    
    \item Through extensive experiments on a diverse array of benchmarks, including all-in-one and single-task restoration scenarios, we demonstrate that DSwinIR consistently achieves competitive or superior performance compared to existing methods on the evaluated datasets.
\end{itemize}

The remainder of this paper is organized as follows. Section~\ref{sec:related_work} provides a comprehensive review of the image restoration literature. Section~\ref{sec:method} elaborates on our proposed method. Section~\ref{sec:experiments} presents extensive experimental validation on a wide array of benchmarks. Finally, Section~\ref{sec:conclusion} concludes the paper by summarizing our key contributions and their broader significance for the field.

\section{Related Work}
\label{sec:related_work}
{Image restoration, a classic and long-standing challenge in computer vision, has seen decades of research~\cite{Digital_IR_Survey}. It has a wide-ranging impact on various downstream tasks, from enhancing visual data for autonomous driving \cite{Almalioglu2022weatherDrive} and improving the robustness of person re-identification \cite{Xu2022Rankinrank} to enabling one-click image correction for cellular segmentation in bioinformatics \cite{Stringer2025Cellpose3}. Nowadays, the advent of deep learning has dramatically accelerated progress in this field \cite{transformer_IR}, leading to remarkable breakthroughs in specialized tasks like denoising\cite{survey_denoise_Michael2023}, dehazing \cite{survey_dehaze_gui2023}, and deraining\cite{survey_derain_chen2023}. More recently, the field's focus has evolved from single-task models to highly versatile \textit{all-in-one frameworks} \cite{survey_all_in_one_jiang2024}, which are more practical for real-world scenarios involving complex, mixed degradations.} In this section, we chart this evolution to contextualize our contribution.

\subsection{Single-Task Image Restoration}
Deep learning, particularly with the rise of Convolutional Neural Networks (CNNs), has long been the cornerstone of image restoration. Leveraging a data-driven paradigm, CNNs have proven highly effective compared to classical approaches \cite{Digital_IR_Survey}, establishing strong foundations in tasks such as image super-resolution \cite{SRCNN}, image denoising \cite{DnCNN}, image deraining \cite{fu2017deraincnn}, and image dehazing \cite{DehazeNet}. Subsequently, numerous more sophisticated studies emerged, as well as those incorporating residual or dense connections \cite{EDSR, RDN}, as well as those exploring channel attention \cite{RCAN} and non-local attention mechanisms \cite{NLSN}. Within specific domains, landmark contributions have guided the field; for instance, in image deraining, Jiang \textit{et al.} proposed MSPFN \cite{jiang2020multi}, which introduced a progressive learning paradigm that inspired a series of subsequent works \cite{Ni2021derain_cvpr, Zhou2021continual_derain, chen2024nerfDerain}. Image dehazing has also seen many representative studies, such as FFANet \cite{Qin2020ffanet}, and more recent explorations with novel network designs \cite{cui2024FocalNet,li2025pami_dehaze}. {Similarly, Feijoo \textit{et al.}~\cite{Feijoo2025DarkIR} exploit a light-weight yet powerful model for low-light image enhancement in the proposed DarkIR.} In parallel, researchers have also focused on developing general restoration architectures applicable to multiple tasks, including {MPRNet} \cite{ZamirA2021Mprnet}, {NAFNet} \cite{chen2022simple}, FSNet \cite{CuiRCK24}, IRNeXT \cite{cui2023irnext}, ConvIR \cite{Cui2024ConvIR}, and MB-TaylorFormer \cite{Jin2025mbtaylor}. However, these powerful models still typically require separate training for each individual task.

Concurrently, to better model the complex, non-local dependencies often present in degradations, Vision Transformers (ViTs) \cite{ViT_Alexey_2021} have garnered significant attention for their powerful dynamic modeling capabilities. While early models like IPT \cite{IPT} were computationally intensive, the window-based local attention of the Swin Transformer \cite{SwinT_LiuL00W0LG21} provided an efficient and effective foundation. This spawned a new generation of influential backbones, including {SwinIR} \cite{SwinIR} and {Uformer} \cite{Uformer}. These advancements were also channeled into powerful task-specific transformers like IDT \cite{Xiao2023IDT} for image deraining and Dehazeformer \cite{Song2023DehazeFormer} for image dehazing. A significant thread of subsequent research has focused extensively on refining the window-based mechanism. For example, {HAT} \cite{HAT_ChenWZ0D23} enlarged window sizes and used overlaps to enhance cross-window communication, while {ELAN} \cite{zhang2022elan} and SRFormer \cite{zhou2023srformer} addressed computational redundancy. More sophisticated attention patterns were also developed, such as the sparse attention in {DRSFormer} \cite{DRSformer}, the strip attention in StripFormer \cite{StripFormer_TsaiPLTL22}, the transposed self-attention in {Restormer} \cite{Restormer}, the integration of an adaptive token dictionary \cite{Zhang2024ATD}, and progressive focused attention \cite{Long2025progressiveSA}. Furthermore, hybrid models like GRL \cite{Li2023GRL} and X-Restormer \cite{Chen2024xRestormer} have explored the fruitful combination of convolutional and attention-based feature extraction.

Alongside these core architectural advancements, several complementary research lines have gained traction. Emerging paradigms like state-space models, exemplified by {MambaIR} \cite{MambaIR,guo2025mambav2}, offer a practical solution for image restoration tasks. {Meanwhile, Shi \textit{et al.}~\cite{Shi2025VmambaIR} introduce the VmambaIR by the proposed omni-scan, and Li \textit{et al.}~\cite{Li2025MaIR} propose the MaIR with the local S-shape scanning and sequential shuffle attention.}
Besides, Wang \textit{et al.} \cite{wang2025navigatingimagerestorationvars} exploited the vision autoregressive model for all-in-one image restoration. Moreover, researchers are exploring other critical facets of image restoration, such as estimating model uncertainty \cite{Ning2021uncertaintydrivenSR, Hong2022uncertaintydehazing,Huang2023uncertaintyDerain,Qiao2024uncertaintyDiffHaze} and leveraging self-supervised pre-training with contrastive learning to derive robust image priors without paired data \cite{Wu2021contrastiveHaze, Zheng2023Curriculardehaze,chen2022unpaircontrastivederain,wu2024modelcontrastive}. These diverse efforts collectively enrich the field and provide a strong foundation for tackling more complex restoration scenarios.

\subsection{All-in-One Image Restoration: The Quest for a Unified Model}
Addressing the challenges of all-in-one image restoration, where a single model must gracefully handle diverse degradations, has inspired a variety of innovative strategies~\cite{survey_all_in_one_jiang2024}. A significant body of work has focused on developing sophisticated prompting mechanisms to guide general-purpose backbones, adapting their behavior to specific tasks. This approach often involves learning adaptive prompts directly from the data. Influential works like {AirNet} \cite{AirNet} and {PromptIR} \cite{PromptIR} demonstrated how a learned degradation representation can effectively modulate a restoration network. To enhance the discriminability of these prompts, subsequent methods introduced explicit regularization, such as the orthogonality constraints in {IDR} \cite{IDR} and the frequency-domain regularization in {AdaIR} \cite{cui2025adair}. {Rajagopalan \textit{et al.}~\cite{Rajagopalan2025AWRaCLe} exploit the task prompt representation with the in-context learning paradigm.}  Others have explored more expressive prompt representations, for instance, by modeling the prompt as a learned distribution as in DPPD \cite{wu2025dppd}. A parallel thread in prompt-based learning injects richer semantic knowledge from large, pre-trained multimodal models. {Zheng \textit{et al.}~\cite{Zeng2025VLUNet} propose the deep unfolding model with pretrained multimodal prior.} Methods like {DA-CLIP} \cite{Luo2024DA-CLIP}, {TextualDegRemoval} \cite{texture}, {Perceive-IR} \cite{zhang2024perceive}, and OmniFuse \cite{Zhang2025onmifusion} leverage models like CLIP \cite{Radford2021CLIP} to provide textual or multimodal guidance. More recently, architectures leveraging Mixture-of-Experts (MoE), such as MoCE-IR \cite{Zamfir2025MOCEIR}, have been exploited to select specialized parameters for different restoration tasks.

{Complementing the work on prompt-based guidance, another significant research direction investigates generative diffusion models \cite{weather_diffusion, DTPM_0001CCXQL024}. For instance, Özdenizci \textit{et al.} \cite{weather_diffusion} proposed the first diffusion-based deweathering model. Following this, Ye \textit{et al.} \cite{DTPM_0001CCXQL024} enhanced the diffusion framework with task-aware adapters to better address multiple degradation tasks.} Besides, some research investigates the optimization challenges inherent in training a single model on diverse data. From a multi-task learning perspective, methods like \cite{wu2024harmony} have explored techniques to mitigate task conflicts, while others like \cite{wu2025debiased} focus on correcting for data and task bias during joint training. Similarly, contrastive learning has been adapted to the all-in-one setting to help the model better decouple degradation-specific features from content-invariant features \cite{tang2025contrastive,wu2025cpl}.

\subsection{Positioning Our Contribution} 
{The power of content-adaptive receptive fields has been clearly established, first bringing significant gains to convolutional networks \cite{DeformableConv_DaiQXLZHW17}. This principle was subsequently adapted for Transformer architectures, such as in the multi-scale deformable attention proposed by Zhu et al. \cite{Zhu2021DeformableDETR} and the deformable attention mechanisms developed by Xia et al. \cite{Xia2022DAT}. Drawing inspiration from this clear line of progress, we introduce the {Deformable Sliding Window (DSwin) Attention}, a new paradigm that is both {token-centric} and {content-aware}.} It directly addresses the core architectural flaws of prior window-based models by eliminating boundary artifacts and enabling content-adaptive receptive fields. By focusing on evolving the foundational architecture, our DSwinIR demonstrates that a sufficiently intelligent and adaptive backbone can achieve state-of-the-art performance, even surpassing prompt-based methods. Our work does not preclude the use of prompts; rather, it provides a significantly more powerful and robust foundation upon which future, more effective prompting strategies can be built.

\begin{figure*}[t] 
\centering 
\includegraphics[width=\textwidth]{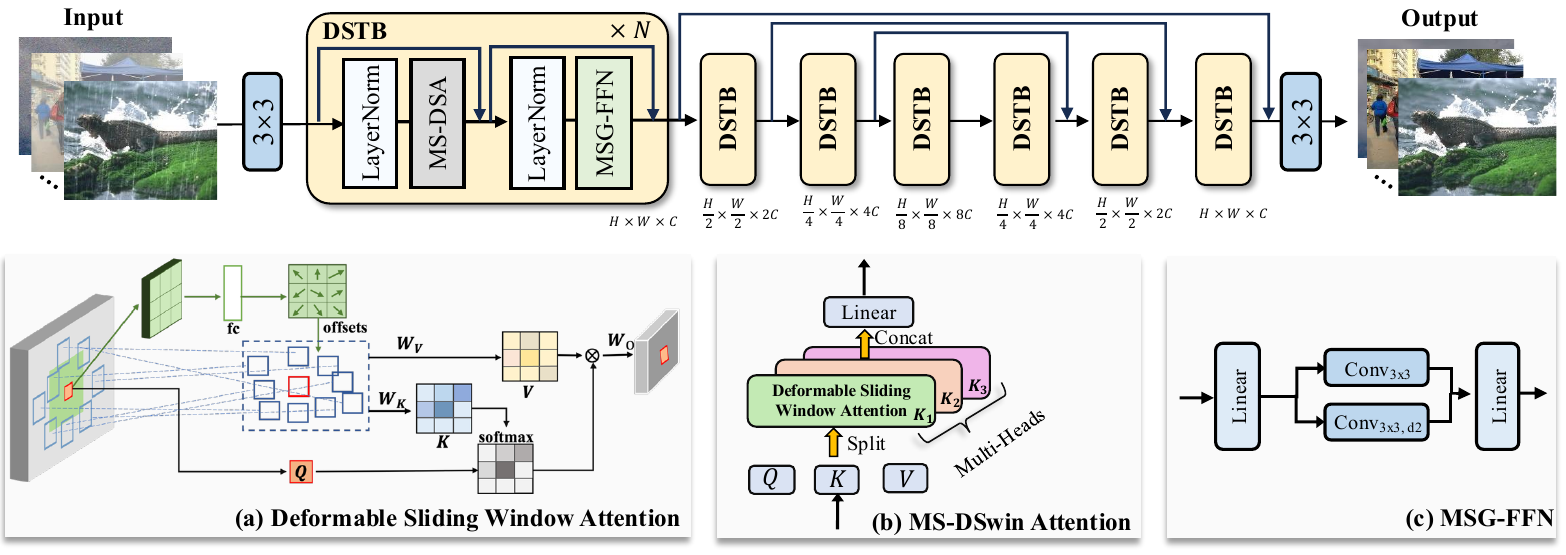} 
\caption{Overview of the proposed DSwinIR architecture. The model is built upon a U-shaped backbone where the core component is the DSwin Transformer Block (DSTB). The key modules include: (a) Deformable Sliding Window Attention (DSwin), which adaptively samples features by learning content-dependent offsets; (b) Multi-Scale Deformable Sliding Window (MS-DSwin) Attention, which integrates multi-scale DSwin attention across multiple attention heads; and (c) Multi-Scale Gated Feed-Forward Network (MSG-FFN), which leverages parallel convolutional branches to enhance feature representation.} 
\label{fig_framework} 
\end{figure*}

\section{Method}
\label{sec:method}
In this section, we introduce DSwinIR. We first outline the overall architecture, then describe the progression from standard window attention to the proposed DSwin attention, including the motivation and formulation. Finally, we present the DSwin Transformer Block and a practical multi‑scale design.

\subsection{Overall Architecture}

We adopt a U‑shaped encoder–decoder architecture with hierarchical feature processing (Fig.~\ref{fig_framework}). The core of this architecture is our novel DSwin Transformer Block, which we will detail in subsequent sections. The entire network is trained end-to-end, optimized with the $L_1$ pixel-wise loss function, which encourages sharp and accurate predictions by penalizing the absolute difference between the restored output and the ground-truth image.

\subsection{From Window Attention to Deformable Sliding Window Attention}
We revisit window‑based attention and develop the Deformable Sliding Window (DSwin) attention from this perspective. In the following subsections, we outline the formulation and design choices.

\subsubsection{Preliminaries: Revisiting Window-based Self-Attention}

We first review standard window‑based self‑attention as depicted in \Cref{fig_intro_dswin}(c). A query token near the center may enjoy a full set of neighbors, but one at the edge—and especially a corner—is starved of context, blind to potentially crucial information just across the arbitrary window boundary. We refer to this as boundary blindness and aim to mitigate it. We briefly formalize its operation here to precisely identify the source of these inherent architectural flaws.

Given a 2D input feature map $\mathbf{X} \in \mathbb{R}^{H \times W \times C}$, standard window-based self-attention first partitions $\mathbf{X}$ into a set of non-overlapping windows $\{\mathbf{X}_w\}$, where each window $\mathbf{X}_w \in \mathbb{R}^{M \times M \times C}$ and $M$ is the window size. Within each window, query ($\mathbf{Q}_w$), key ($\mathbf{K}_w$), and value ($\mathbf{V}_w$) projections are computed. The self-attention is then calculated as:
\begin{equation}
\mathrm{Attention}(\mathbf{Q}_w, \mathbf{K}_w, \mathbf{V}_w) = \mathrm{Softmax}\left(\frac{\mathbf{Q}_w \mathbf{K}_w^T}{\sqrt{d}} + \mathbf{B}\right)\mathbf{V}_w,
\end{equation}
where $d$ is the head dimension and $\mathbf{B}$ is a learnable relative position bias. As discussed in the introduction, this formulation, while efficient, inherently suffers from insufficient cross-window interaction. 

\subsubsection{Token‑Centric Sliding Window}
Boundary artifacts largely stem from non‑overlapping partitions. We replace window‑centric partitioning with a \emph{token‑centric} sliding window, inspired by sliding convolution.

For a given query token at position $(i,j)$, we define its local neighborhood of size $k \times k$. This operation extracts a feature patch $\mathcal{N}_{k}(i,j) \in \mathbb{R}^{k \times k \times C}$ centered at each position $(i,j)$. It effectively transforms the feature map $\mathbf{X}$ into a sequence of patches, where each patch contains the keys and values for a corresponding query token. The attention for the query token $\mathbf{q}_{i,j}$ is then computed over the keys $\mathbf{K}_{i,j}$ and values $\mathbf{V}_{i,j}$ extracted from its neighborhood $\mathcal{N}_{k}(i,j)$:
\begin{equation}
\mathbf{Y}_{i,j}^{\text{slide}} = \sum_{(u,v) \in \mathcal{N}_k} \mathrm{Softmax}_{uv}\left(\frac{\mathbf{q}_{i,j} \cdot \mathbf{k}_{i+u, j+v}}{\sqrt{d}}\right) \mathbf{v}_{i+u, j+v},
\end{equation}
where $(u,v)$ are the relative coordinates within the $k \times k$ neighborhood. Unlike non‑overlapping windows, this sliding mechanism reduces the information bottleneck at boundaries and facilitates communication across spatial locations.

\subsubsection{Content‑Aware Deformable Sampling}
Sliding windows alleviate boundary issues but still rely on a fixed, rectangular sampling grid. This fixed partitioning is misaligned with the arbitrary geometry of natural image structures and degradation patterns, such as the fine, diagonal lines of rain streaks or the amorphous, non-uniform nature of haze. Such a mismatch compromises the model's ability to efficiently aggregate contextually relevant features, pointing to a critical need for a more flexible mechanism that can dynamically shape its receptive field in a content-aware manner.

\paragraph{Offset Prediction and Differentiable Sampling} For each query token $\mathbf{q}_{i,j}$, we predict a set of 2D spatial offsets $\{\Delta\mathbf{p}_{i,j}^{(u,v)}\}$ for all $k \times k$ sampling points in its neighborhood. These offsets are generated by a lightweight offset prediction network $f_{\theta}$ that takes the query features as input:
\begin{equation}
\{\Delta\mathbf{p}_{i,j}^{(u,v)}\} = f_{\theta}(\mathbf{q}_{i,j}).
\end{equation}

The network $f_{\theta}$ is implemented as a lightweight module consisting of a depth-wise separable convolution, followed by a GELU activation function, and a final linear projection. No explicit offset regularization was required in our experiments; offsets are learned end‑to‑end via the restoration loss. The deformed key and value features are then sampled from the original feature map at the new coordinates $(i+u+\Delta u_{i,j}^{(u,v)}, j+v+\Delta v_{i,j}^{(u,v)})$. Since the offsets are fractional, we use \textit{differentiable bilinear interpolation} to sample the features, ensuring the entire process is end-to-end trainable. A value at a fractional location $(x,y)$ is computed as:
\begin{equation}
\mathbf{F}(x,y) = \sum_{i',j'} w(i',j') \cdot \mathbf{X}( \lfloor x \rfloor+i', \lfloor y \rfloor+j' ),
\end{equation}
where $(i', j') \in \{(0,0), (0,1), (1,0), (1,1)\}$ are the four integer grid neighbors, and $w(i',j')$ are the bilinear interpolation weights, which are proportional to the distance from $(x,y)$ to the opposite corner.

\paragraph{Deformable Sliding Window Attention} By combining these elements, the final Deformable Sliding Window (DSwin) Attention output for a token at position $(i,j)$ is formulated as:
\begin{equation}
\mathbf{Y}_{i,j} = \sum_{(u,v) \in \mathcal{N}_k} \alpha_{i,j}^{(u,v)} \cdot \mathbf{v}_{\text{interp}}(i+u+\Delta u_{i,j}^{(u,v)}, j+v+\Delta v_{i,j}^{(u,v)}),
\label{eq:dswin_final}
\end{equation}
where $\alpha_{i,j}^{(u,v)}$ represents the attention weights computed using the deformed keys, and $\mathbf{v}_{\text{interp}}$ denotes the value sampled via bilinear interpolation. This formulation combines sliding and deformable sampling: sliding provides overlapping local context; deformable offsets adapt sampling to content, as illustrated in \Cref{fig_intro_dswin}(d).

\begin{figure}[!t] 
\centering 
\includegraphics[width=\linewidth]{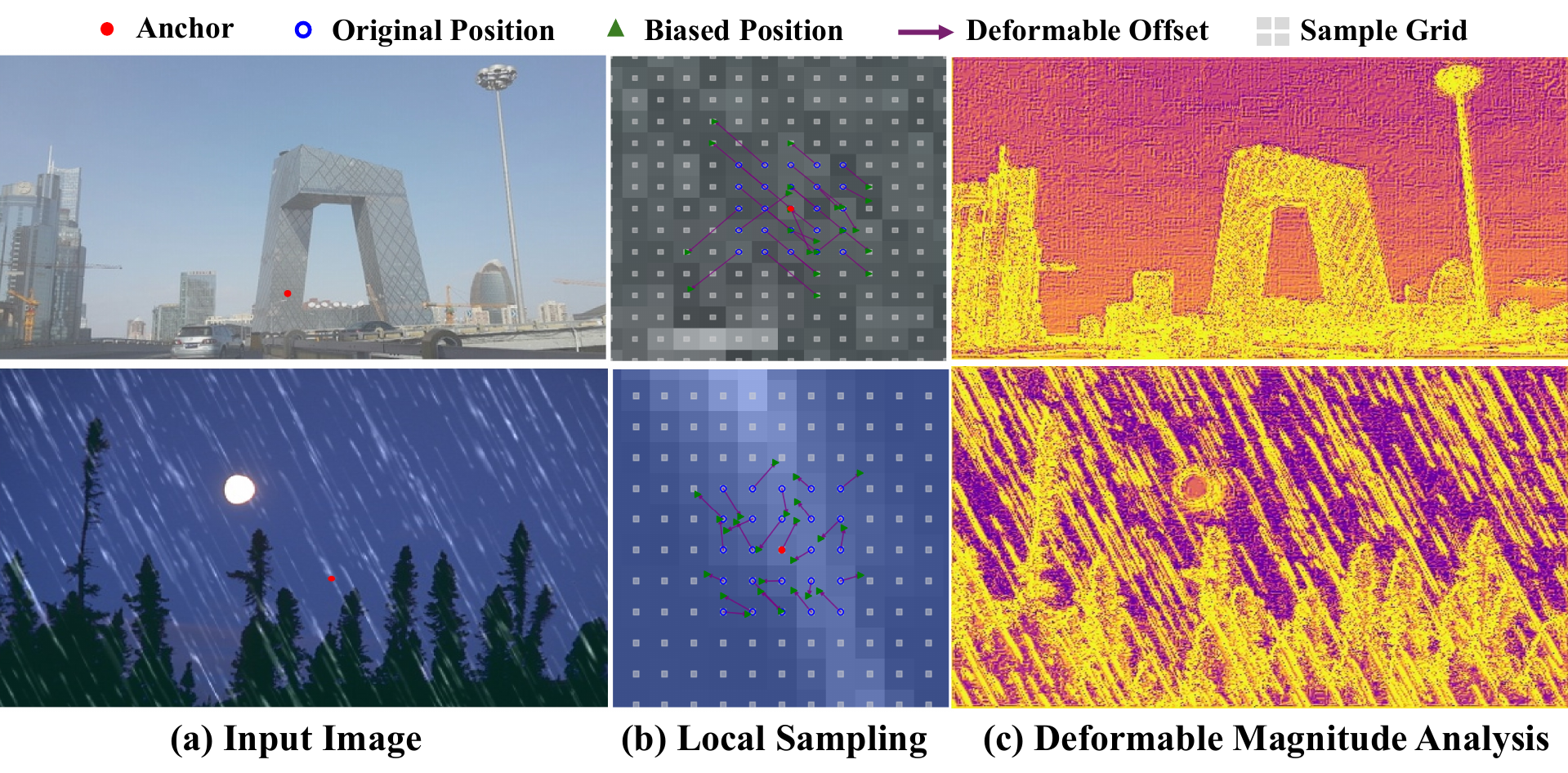} 
\caption{{Visualization of the content-aware sampling mechanism in DSwinIR.
The figure presents a qualitative analysis of our proposed DSwin Attention. (a) The input image, with a red dot indicating the anchor location for local analysis. (b) The corresponding local sampling pattern, showing the reference grid, the adaptively sampled points, and the learned offsets. (c) The Deformation Magnitude Map, which visualizes the offset distance ($d = \sqrt{dx^2 + dy^2}$) as a heatmap, where brighter intensity signifies a larger offset.}} 
\label{fig:offset_viz} 
\end{figure}

\begin{table*}[!t]
\renewcommand\arraystretch{1.75} 
\centering
\caption{Comprehensive overview of the experimental settings used to evaluate DSwinIR. Each setting is defined by its restoration tasks, degradation types, and corresponding datasets for training and testing.}
\label{tab_tasks_expanded}
\resizebox{\textwidth}{!}{%
\begin{tabular}{l|c|l|l}
\toprule[1pt]
\textbf{Experiment Settings} & \textbf{Tasks} & \textbf{Degradation Types} & \textbf{Datasets} \\
\hline

Three-Task Restoration & 3 & Denoising, Dehazing, Deraining & \makecell[l]{BSD400 \cite{arbelaez2011contour}, WED \cite{ma2017waterloo}, RESIDE (SOTS) \cite{li2019benchmarking}, Rain100L \cite{yang2017deep}, BSD68 \cite{martin2001database1}} \\
\hline

Five-Task Restoration & 5 & \makecell[l]{Adds Motion Deblurring \& Low-Light \\ Enhancement to the three-task setup} & \makecell[l]{Adds GoPro \cite{nah2017deep} and LOLv1 \cite{wei2018deep} to the three-task training set} \\
\hline

Synthetic Deweathering & 3 & Snow, Raindrop, Outdoor Rain & Snow100K \cite{Snow100K_dataset}, Raindrop \cite{raindrop}, Outdoor-Rain \cite{test1_dataset_allweather} \\
\hline
Real-World Deweathering & 3 & Real Rain, Real Snow, Real Haze & SPA-Data \cite{SPANet_WangY0C0L19}, RealSnow\cite{weather_data}, REVIDE\cite{Zhang2021Realhaze}  \\
\hline
{Real-World Deweathering} & {3} & {All-Time Adverse Weather Types} & {WeatherBench~\cite{Guan2025WeatherBench}}  \\
\hline

{Composite Degradation} & {11} & \makecell[l]{{Low-Light, Haze, Rain, Snow,} \\ {and their combinations}} & {DIV2K \cite{Agustsson_2017_CVPR_Workshops} with corresponding degradations} \\
\bottomrule[1pt]
\end{tabular}%
}
\end{table*}
\subsection{DSwin Transformer Block}
The DSwin Transformer Block integrates DSwin attention and a feed‑forward network with a complementary multi‑scale strategy to strengthen feature extraction.

\subsubsection{Multi-Scale DSwin (MS-DSwin) Attention}
Image content and degradations manifest across a spectrum of scales. The standard multi-head attention framework provides a natural and computationally efficient vehicle for capturing such multi-scale information. Instead of treating all attention heads identically, we leverage this structure to explicitly assign different spatial extents to different groups of heads, as depicted in \Cref{fig_framework}(b). Specifically, for a model with $N$ heads, we partition them into groups, with each group $h$ being assigned a unique sliding window kernel size $k_h$. This allows the model to simultaneously aggregate features from both compact local neighborhoods and broader contextual regions within a single MS-DSwin module. Outputs from all heads are concatenated and projected to fuse multi‑scale information at a small additional cost.

\subsubsection{Multi-Scale Gated Feed-Forward Network (MSG-FFN)}
A standard feed‑forward network (FFN) processes tokens independently via point‑wise linear layers, without explicit spatial context. To keep a multi‑scale design throughout the block, we introduce a Multi‑Scale Gated FFN (MSG‑FFN): after channel expansion, features split into parallel depth‑wise convolution branches with different kernel sizes or dilations, followed by gating and projection. As shown in \Cref{fig_framework}(c), after an initial channel expansion layer, the MSG-FFN splits the features into parallel branches. Each branch employs a depth-wise convolution with a different kernel size or dilation rate, explicitly re-introducing spatial reasoning at multiple scales.

\subsection{Remarks}
It is crucial to position our contribution not as another incremental improvement, but as a complementary perspective in designing attention mechanisms for dense prediction tasks. Previous works typically resort to hand-crafted patterns. Our DSwin Attention, in contrast, introduces the token-centric and content-aware paradigm. This addresses the core issues of \textit{cross-window interaction} and \textit{inflexible receptive fields} directly and adaptively.

\textit{Content-Aware Property:} To provide intuition for the dynamic behavior of our proposed DSwin Attention, we present a visual analysis of its learned offsets in \Cref{fig:offset_viz}. This visualization highlights two key properties of our module. First, at a micro-level (middle column), the learned offsets enable the sampling points to break free from the rigid reference grid, allowing the module to adaptively fetch features from more informative, non-adjacent locations. Second, at a macro-level (right column), the module exhibits a strong content-aware capability. We visualize the offset magnitude ($d = \sqrt{dx^2 + dy^2}$) as a heatmap and observe a clear correlation with input complexity. The module learns significantly larger offsets for challenging regions (e.g., complex textures or heavy degradation), indicating an active search for relevant information over a wider area. Conversely, it reverts to a local operation (minimal offsets) in simple, uniform regions. This behavior validates that our DSwin Attention operates as a dynamic, content-aware operator rather than a static receptive field expander. We provide a more quantitative analysis of the resulting receptive field in our ablation study.

\begin{table*}[!htb]
\setlength{\abovecaptionskip}{0pt}
  \caption{Quantitative comparisons for \textit{Setting 1: three distinct degradation tasks}. * denotes results reported by prior work \cite{PromptIR,zhang2024perceive}.} 
  \label{tab_three_task}
  \centering
    \renewcommand\arraystretch{1.2}
    \resizebox{\linewidth}{!}{
\begin{tabular}{c|l|l|ccc|c|c|c}
    \toprule[1pt]

 \multirow{2}{*}{\textbf{Type}}& \multirow{2}{*}{\textbf{Method}} &  \multirow{2}{*}{\textbf{Venue}} 
    & \multicolumn{3}{c|}{\textbf{Denoising} (CBSD68\cite{martin2001database1})}
    & \multicolumn{1}{c|}{\textbf{Dehazing}}
    & \multicolumn{1}{c|}{\textbf{Deraining}}
    & \multirow{2}{*}{\textbf{Average}}
    \\
    \cline{4-8}
   & &  & $\sigma = 15$ & $\sigma = 25$ & $\sigma = 50$  
    & SOTS \cite{li2019benchmarking}
    & Rain100L \cite{yang2017deep}
    &  \\
    \hline

  \multirow{9}{*}{\rotatebox{90}{\textit{General}}} 
  & MPRNet \cite{zamir2021multi} & CVPR'21
    & 33.27/0.920  & 30.76/0.871  & 27.29/0.761    & 28.00/0.958   & 33.86/0.958  & 30.63/0.894 \\
   & Uformer~\cite{Uformer} & CVPR'22
     & 33.55/0.922 
     & 30.71/0.859 
     & 27.63/0.786
     & 27.13/0.953 
     & 34.28/0.961 
     & 30.66/0.896 \\
   & Restormer \cite{Restormer} & CVPR'22 
    & 33.72/0.930  & 30.67/0.865  & 27.63/0.792    & 27.78/0.958   & 33.78/0.958  & 30.75/0.901 \\
  & NAFNet \cite{chen2022simple} & ECCV'22 
    & 33.03/0.918  & 30.47/0.865  & 27.12/0.754    & 24.11/0.928   & 33.64/0.956  & 29.67/0.844  \\
   & FSNet* \cite{CuiRCK24}  &  TPAMI'23 
    & 33.81/0.930   & 30.84/0.872   & 27.69/0.792     & 29.14/0.968    & 35.61/0.969   & 31.42/0.906   \\
    & DRSformer* \cite{DRSformer} & CVPR'23 
    &33.28/0.921   &30.55/0.862   &27.58/0.786     &29.02/0.968    & 35.89/0.970   &31.26/0.902 \\
   & MambaIR* \cite{MambaIR} & ECCV'24
    & 33.88/0.931  & 30.95/0.874  & 27.74/0.793    &29.57/0.970   & 35.42/0.969 & 31.51/0.907 \\  
   & VmambaIR~\cite{Shi2025VmambaIR} & TCSVT'25
     & 33.75/0.927 
     & 31.10/0.879 
     & 27.81/0.782
     & 29.75/0.970 
     & 35.63/0.970 
     & 31.61/0.906 \\
   & MaIR~\cite{Li2025MaIR} & CVPR'25
     & 33.60/0.925 
     & 30.98/0.876 
     & 27.69/0.779
     & 28.94/0.965 
     & 34.52/0.964 
     & 31.15/0.902 \\
    
    \hline

  \multirow{13}{*}{\rotatebox{90}{\textit{All-in-One}}} 
  &  DL \cite{dl} & TPAMI'19
    & 33.05/0.914  & 30.41/0.861  & 26.90/0.740    & 26.92/0.391   & 32.62/0.931  & 29.98/0.875 \\
   & AirNet \cite{AirNet} & CVPR'22 
    & 33.92/0.932  & 31.26/0.888  & 28.00/0.797    & 27.94/0.962   & 34.90/0.967  
    & 31.20/0.910  \\
   & IDR* \cite{IDR}  & CVPR'23 
    & 33.89/0.931  & 31.32/0.884  & 28.04/0.798    & 29.87/0.970   & 36.03/0.971  & 31.83/0.911  \\
  & PromptIR \cite{PromptIR}  & NeurIPS'23
    & 33.98/0.933  & 31.31/0.888  & 28.06/0.799    
    & 30.58/0.974   & 36.37/0.972  
    & 32.06/0.913  \\
  & Gridformer* \cite{Gridformer}  &  IJCV'24
    & 33.93/0.931  & 31.37/0.887  & 28.11/0.801    
    & 30.37/0.970   & 37.15/0.972  
    & 32.19/0.912  \\
  & NDR \cite{NDR}  & TIP'24 
    & 34.01/0.932  & 31.36/0.887  & 28.10/0.798    & 28.64/0.962   
    & 35.42/0.969  & 31.51/0.910 \\
 & InstructIR \cite{InstructIR}  & ECCV'24 
    & \textbf{34.15}/0.933  & 31.52/\underline{0.890}  & \underline{28.30}/\textbf{0.804} 
    & 30.22/0.959   & 37.98/0.978  
    & 32.43/0.913 \\
  & TextualDegRemoval\cite{texture} & CVPR'24
    & 34.01/0.933 & 31.39/\underline{0.890}  
    & 28.18/0.802  & \underline{31.63}/\textbf{0.980}   
    & 37.58/0.979  & 32.63/0.917 \\
  & AdaIR \cite{cui2025adair} & ICLR'25
    & 34.12/\textbf{0.935} 
    & 31.45/\textbf{0.892} 
    & 28.19/0.802 
    & 31.06/\textbf{0.980} 
    & \underline{38.64}/\underline{0.983}
    & 32.69/\underline{0.918} \\
  &Perceive-IR \cite{zhang2024perceive} & TIP'25
    & \underline{34.13}/\underline{0.934} 
    & \underline{31.53}/\underline{0.890}  
    & \textbf{28.31}/\textbf{0.804}  
    & \underline{30.87}/\underline{0.975}   
    & 38.29/0.980  
    & 32.63/0.917 \\
  & PoolNet~\cite{PoolNet} & TIP'25
     & 34.10/\textbf{0.935} 
     & 31.45/\textbf{0.892}
     & 28.18/\underline{0.803}
     & 30.94/\textbf{0.980} 
     & \underline{38.54}/\underline{0.983} 
     & 32.64/\textbf{0.919} \\
  & VLU-Net~\cite{Zeng2025VLUNet} & CVPR'25
     & \underline{34.13}/\textbf{0.935} 
     & 31.48/\textbf{0.892} 
     & 28.23/\textbf{0.804}
     & 30.71/\textbf{0.980} 
     & \textbf{38.93}/\textbf{0.984} 
     & \underline{32.70}/\textbf{0.919} \\
   &   \cellcolor{my_color}\textbf{DSwinIR (Ours)} &  \cellcolor{my_color}2025  
    & \cellcolor{my_color}34.12/0.933 
    & \cellcolor{my_color}\textbf{31.59}/\underline{0.890}  
    & \cellcolor{my_color}\textbf{28.31}/\underline{0.803}  
    & \cellcolor{my_color}\textbf{31.86}/\textbf{0.980}   
    & \cellcolor{my_color}37.73/\underline{0.983}  
    & \cellcolor{my_color}\textbf{32.72}/0.917
 \\
    
  \bottomrule[1pt]
\end{tabular}

  }
\end{table*}
\begin{figure*}[!h] 
\centering \includegraphics[width=\linewidth]{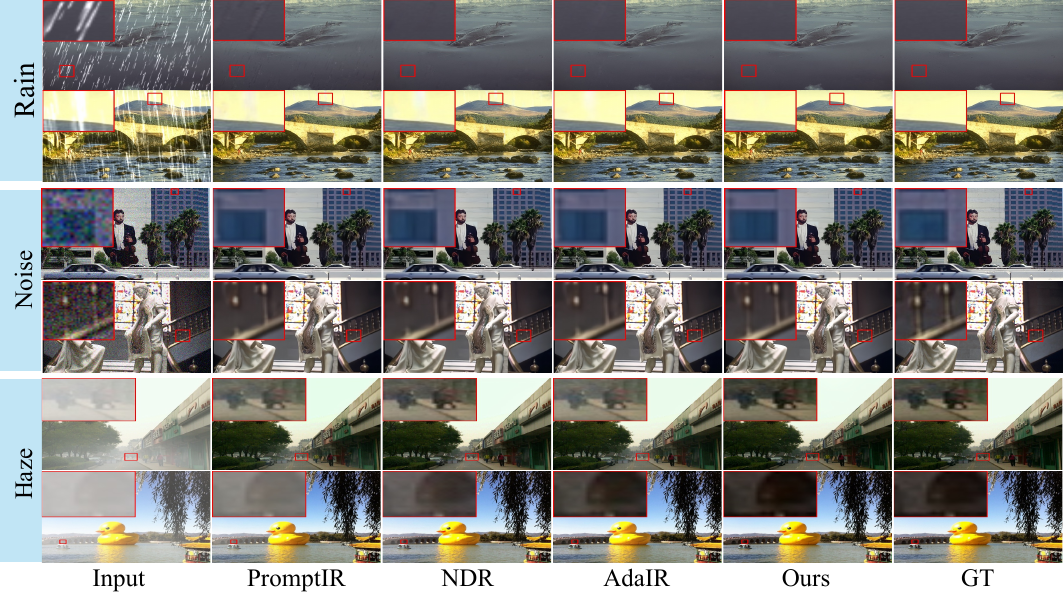} 
\caption{Visual comparison of restoration results across three  degradation tasks: noise removal (top row), rain streak removal (middle row), and dehazing (bottom row). Zoom-in regions (shown in colored boxes) demonstrate that our method achieves superior detail preservation and degradation removal.} 
\label{fig_3_task} 
\end{figure*}

\section{Experiments}
\label{sec:experiments}

In this section, we conduct a comprehensive set of experiments to rigorously evaluate the performance of our proposed DSwinIR. We first detail the experimental setup, including datasets, evaluation protocols, and implementation specifics. We then present extensive quantitative and qualitative comparisons against state-of-the-art methods across a wide range of image restoration tasks. Finally, we provide in-depth ablation studies to dissect the contribution of each component of our model and analyze its underlying mechanisms.
\subsection{Experimental Setup}
To comprehensively validate the capabilities of DSwinIR, we have designed a multi-faceted evaluation protocol spanning a wide range of restoration scenarios, from standard multi-task benchmarks to complex and composite degradations. This section details the benchmarks, competing methods, and implementation specifics of our experiments.

\begin{table*}[tb]
\setlength{\abovecaptionskip}{0pt}
  \caption{Quantitative comparisons for \textit{Setting 2: five distinct degradation tasks}. * indicates numbers reported by prior work \cite{PromptIR,zhang2024perceive}.}
  \label{tab_five_task}
  \centering
  \small
  \renewcommand\arraystretch{1.2}
    \resizebox{\linewidth}{!}{

\begin{tabular}{c|l|l|c|c|c|c|c|c}
\toprule[1pt]
\multirow{2}{*}{\bf Type} & \multirow{2}{*}{\bf Method} & \multirow{2}{*}{\bf Venue} 
& \multicolumn{1}{c|}{\bf Denoising}
& \multicolumn{1}{c|}{\bf Dehazing}
& \multicolumn{1}{c|}{\bf Deraining}
& \multicolumn{1}{c|}{\bf Deblurring}
& \multicolumn{1}{c|}{\bf Low-Light}
& \multirow{2}{*}{\bf Average}
\\ \cline{4-8} 
&  & & CBSD68 \cite{martin2001database1} & SOTS \cite{li2019benchmarking} & Rain100L \cite{yang2017deep} & GoPro \cite{nah2017deep} & LOL \cite{wei2018deep} & \\
\hline

\multirow{11}{*}{\rotatebox{90}{\textit{General}}} 
& SwinIR \cite{SwinIR} & ICCVW'21 
& 30.59/0.868  & 21.50/0.891  & 30.78/0.923    & 24.52/0.773   & 17.81/0.723  & 25.04/0.835 \\
& Uformer~\cite{Uformer} & CVPR'22
& 30.49/0.883 & 23.40/0.919 & 33.32/0.962 & 26.26/0.807 & 21.16/0.817 & 26.93/0.877\\
& MIRNet-v2 \cite{MIRNet_v2} & TPAMI'22  
& 30.97/0.881  & 24.03/0.927  & 33.89/0.954    & 26.30/0.799   & 21.52/0.815  & 27.34/0.875\\
& Restormer \cite{Restormer} & CVPR'22 
& \underline{31.49}/0.884  & 24.09/0.927  & 34.81/0.962    & 27.22/0.829   & 20.41/0.806  & 27.60/0.881\\
& NAFNet \cite{chen2022simple} &  ECCV'22 
& 31.02/0.883  & 25.23/0.939  & 35.56/0.967    & 26.53/0.808   & 20.49/0.809  & 27.76/0.881\\
& DRSformer* \cite{DRSformer} &  CVPR'23 
& 30.97/0.881   & 24.66/0.931   & 33.45/0.953     & 25.56/0.780    & 21.77/0.821   & 27.28/0.873\\
& Retinexformer* \cite{Retinexformer}  & ICCV'23  
& 30.84/0.880  & 24.81/0.933   & 32.68/0.940     & 25.09/0.779    & 22.76/0.834   & 27.24/0.873\\
& FSNet* \cite{CuiRCK24} &  TPAMI'23  
& 31.33/0.883  & 25.53/0.943  & 36.07/0.968    & 28.32/0.869   & 22.29/0.829  & 28.71/0.898\\
& MambaIR* \cite{MambaIR} & ECCV'24 
& 31.41/0.884  & 25.81/0.944  & 36.55/0.971    & 28.61/0.875   & 22.49/0.832  & 28.97/0.901\\
& VMambaIR~\cite{Shi2025VmambaIR} & TCSVT'25 
& 30.81/0.879 & 27.65/0.949 & 34.26/0.960 & 26.02/0.799 & 22.38/0.827 & 28.22/0.883\\
& MaIR~\cite{Li2025MaIR} & CVPR'25
& 30.98/0.882 & 27.72/0.950 & 34.04/0.946 & 25.68/0.791 & 22.17/0.821 & 28.11/0.876\\

\hline

\multirow{13}{*}{\rotatebox{90}{\textit{All-in-One}}} 
& DL \cite{dl} & TPAMI'19 
& 23.09/0.745  & 20.54/0.826  & 21.96/0.762    & 19.86/0.672   & 19.83/0.712  & 21.05/0.743\\
& TAPE \cite{TAPE} & ECCV'22 
& 30.18/0.855  & 22.16/0.861  & 29.67/0.904    & 24.47/0.763   & 18.97/0.621  & 25.09/0.801\\
& Transweather \cite{Transweather} &  CVPR'22 
& 29.00/0.841  & 21.32/0.885  & 29.43/0.905    & 25.12/0.757   & 21.21/0.792  & 25.22/0.836\\
& AirNet \cite{AirNet} &   CVPR'22 
& 30.91/0.882  & 21.04/0.884  & 32.98/0.951    & 24.35/0.781   & 18.18/0.735  & 25.49/0.846\\
& IDR \cite{IDR} &  CVPR'23 
& \textbf{31.60}/0.887  & 25.24/0.943  & 35.63/0.965    & 27.87/0.846   & 21.34/0.826  & 28.34/0.893\\
& PromptIR* \cite{PromptIR} &  NeurIPS'23 
& 31.47/0.886  & 26.54/0.949  & 36.37/0.970    & 28.71/\textbf{0.881}   & 22.68/0.832  & 29.15/0.904\\
& Gridformer* \cite{Gridformer} &  IJCV'24 
& 31.45/0.885  & 26.79/0.951  & 36.61/0.971 & \underline{29.22}/\underline{0.884} & 22.59/0.831 & 29.33/0.904\\
& InstructIR \cite{InstructIR} &  ECCV'24 
& 31.40/\textbf{0.887}  & 27.10/0.956  & 36.84/0.973    
& \underline{29.40}/\textbf{0.886}   & \textbf{23.00}/0.836  
& 29.55/0.907\\
& AdaIR\cite{cui2025adair} & ICLR'25
& 31.35/\underline{0.889} 
& \underline{30.53}/\underline{0.978}
& \underline{38.02}/\underline{0.981} 
& 28.12/0.858 
& \textbf{23.00}/\textbf{0.845} 
& \textbf{30.20}/\underline{0.910}\\
& Perceive-IR \cite{zhang2024perceive} & TIP'25
& 31.44/\textbf{0.887} 
& 28.19/0.964  
& 37.25/0.977   
& \textbf{29.46}/\textbf{0.886}  
& \underline{22.88}/0.833  
& 29.84/0.909\\
& PoolNet~\cite{PoolNet} & TIP'25
& 31.24/\textbf{0.887} & 30.25/0.977 & 37.85/\underline{0.981} & 27.66/0.844 & 22.66/0.841 & 29.93/0.906\\
& VLU-Net~\cite{Zeng2025VLUNet} & CVPR'25
& 31.43/\textbf{0.891} & \textbf{30.84}/\textbf{0.980} & \textbf{38.54}/\textbf{0.982}  & 27.46/0.840 & 22.29/0.833 & 30.11/0.905\\

& \cellcolor{my_color}\textbf{DSwinIR (Ours)} & \cellcolor{my_color}2025
& \cellcolor{my_color}31.34/0.885  
& \cellcolor{my_color}30.09/0.975  
& \cellcolor{my_color}37.77/\textbf{0.982}    
& \cellcolor{my_color}29.17/0.879  
& \cellcolor{my_color}22.64/\underline{0.843}  
& \cellcolor{my_color}\textbf{30.20}/\textbf{0.913}\\

\bottomrule[1pt]
\end{tabular}
  
  }
\end{table*}

\subsubsection{Evaluation Benchmarks and Metrics}

Our evaluation is structured across four distinct experimental settings, each designed to probe a specific aspect of our model's performance. For all quantitative evaluations, we employ the widely-recognized Peak Signal-to-Noise Ratio (PSNR) and Structural Similarity Index Measure (SSIM). Higher values in both metrics indicate superior restoration quality.

\paragraph{Multi-Task All-in-One Image Restoration}
This setting assesses the ability of a single model to serve as a versatile, general-purpose restorer. Following established conventions, we evaluate on two standard benchmarks. The \textit{three-task} benchmark \cite{AirNet, PromptIR} comprises Gaussian denoising, deraining, and dehazing. The more challenging \textit{five-task} benchmark \cite{cui2025adair} extends this by incorporating motion deblurring and low-light enhancement, demanding greater generalization from a single set of model weights. The specific datasets for training and testing are detailed in \Cref{tab_tasks_expanded}.

\paragraph{Adverse Weather Removal}
This setting focuses on the practical and challenging domain of deweathering. We test DSwinIR's robustness on both synthetic and real-world data. For the synthetic evaluation, we use the ``AllWeather'' collection \cite{test1_dataset_allweather}, which includes images corrupted by snow, raindrops, and heavy rain, allowing for a controlled assessment across varied weather phenomena. To validate real-world efficacy, we employ a challenging benchmark \cite{weather_data} composed of datasets with authentic rain (SPA-Data \cite{SPANet_WangY0C0L19}), snow (RealSnow \cite{weather_data}), and haze (REVIDE \cite{Zhang2021Realhaze}), which feature complex, non-uniform artifacts that are the true test of a model's generalization power. {To further validate our method, we also conduct evaluations on the recent WeatherBench~\cite{Guan2025WeatherBench} dataset. This large-scale, real-world benchmark provides 41,402 training and 600 testing pairs. It presents a significant challenge by incorporating a diverse array of adverse weather conditions, including foggy, raining, and snowing scenarios captured during both daytime and nighttime.}

\paragraph{Composite Degradation}
To push the boundaries of evaluation, we evaluate a composite degradation benchmark introduced in \cite{guo2024onerestore} to assess the model's performance in more realistic and complex scenarios where multiple corruptions coexist. This benchmark is crucial for testing a model's ability to disentangle and simultaneously mitigate interacting artifacts. We adopt a challenging test set by combining four degradation types: low-light (l), haze (h), rain (r), and snow (s). The evaluation protocol includes all single degradations, all six dual-degradation combinations (e.g., l+h, h+r), and two triple-degradation combinations (l+h+r, l+h+s). Success in this setting signals a higher level of architectural intelligence and robustness.

\paragraph{Single-Task Specialization}
While all-in-one models demonstrate versatility, a truly superior architecture must also excel when specialized. This experimental setting investigates the ``specialization power" of DSwinIR by training it exclusively for a single task. We conduct focused evaluations on image deraining (Rain100L and the real-world SPA-Data) and dehazing (SOTS). {In addition, we employ further low-light evaluation benchmarks, including the real-world LOLv2 dataset. We also evaluate DSwinIR on a mixed task involving both low-light enhancement and deblurring on the LowBlur dataset, following the protocol from~\cite{Feijoo2025DarkIR}.} Outperforming state-of-the-art specialist models in this context provides evidence of our architecture's significant strength.

\begin{table*}[!t]
\setlength{\abovecaptionskip}{5pt}
\setlength{\belowcaptionskip}{0pt}
\small
\renewcommand\arraystretch{1.2} 
    \centering
    \caption{{Quantitative comparisons for {adverse weather removal}} including rain, haze, and snow.  Missing values are denoted by `---'.}
    \label{tab_allweather}
      \setlength{\tabcolsep}{4pt}
\scalebox{1}{
\begin{tabular}{cllcccccccccc}
    \toprule[1pt]
    \multirow{2}{*}{\textbf{Type}} & \multirow{2}{*}{\textbf{Methods}} &  \multirow{2}{*}{Venue}   &  \multicolumn{2}{c}{\textbf{Snow100K-S}~\cite{Snow100K_dataset}} & \multicolumn{2}{c}{\textbf{Snow100K-L}~\cite{Snow100K_dataset}} & \multicolumn{2}{c}{\textbf{Outdoor-Rain}~\cite{test1_dataset_allweather}} & \multicolumn{2}{c}{\textbf{RainDrop}~\cite{raindrop}} & \multicolumn{2}{c}{Average} \\
     
    \cmidrule(r){4-5} \cmidrule(r){6-7} \cmidrule(r){8-9} \cmidrule(r){10-11} \cmidrule{12-13}
    & & & PSNR & SSIM & PSNR & SSIM & PSNR & SSIM & PSNR & SSIM & PSNR & SSIM \\
        \midrule
   \multirow{8}{*}{\rotatebox{90}{\textit{Task-Specific}}} 
   &  SPANet~\cite{SPANet_WangY0C0L19}    & CVPR'19    & 29.92 & 0.8260 & 23.70 & 0.7930 & -- & -- & -- & -- & -- & -- \\
     &   DesnowNet~\cite{Snow100K_dataset}     & TIP'18  & 32.33 & 0.9500 & 27.17 & 0.8983 & -- & -- & -- & -- & -- & -- \\

     & HRGAN~\cite{outdoor}      &    CVPR'19  & -- & -- & -- & -- & 21.56 & 0.8550 & -- & -- & -- & -- \\
     &   MPRNet~\cite{zamir2021multi} &  CVPR'21 &     -- & -- & -- & -- & 28.03 & 0.9192 & -- & -- & -- & -- \\
     &  AttentiveGAN~\cite{raindrop}      & CVPR'18 & -- & -- & -- & -- & -- & -- & 31.59 & 0.9170 & -- & -- \\
     &   IDT~\cite{Xiao2023IDT} & TIP'22  &     -- & -- & -- & -- & -- & -- & 31.87 & 0.9313 & -- & -- \\
     &   NAFNet~\cite{chen2022simple} &   ECCV'22  &  34.79 & 0.9497 & 30.06 & 0.9017 & 29.59 & 0.9027 & -- & -- & -- & -- \\
     &  Restormer~\cite{Restormer}      & CVPR'22 & 36.01 & 0.9579 & 30.36 & 0.9068 & 30.03 & 0.9215 & 32.18 & 0.9408 & -- & -- \\
        \midrule
    \multirow{12}{*}{\rotatebox{90}{\textit{All-in-One}}} 
     & All-in-One~\cite{as2020}  & CVPR'20  & -- & -- & 28.33 & 0.8820 & 24.71 & 0.8980 & 31.12 & 0.9268 & 28.05 & 0.9023 \\
     &   TransWeather~\cite{Transweather}   & CVPR'22   & 32.51 & 0.9341 & 29.31 & 0.8879 & 28.83 & 0.9000 & 30.17 & 0.9157 & 30.20 & 0.9094 \\
     &  Chen \textit{et al.}~\cite{ChenHTYDK22} &  CVPR'22  &   34.42 & 0.9469 & 30.22 & 0.9071 & 29.27 & 0.9147 & 31.81 & 0.9309 & 31.43 & 0.9249 \\
     &  WGWSNet~\cite{ZhuWFYGDQH23}    &  CVPR'22  & 34.31 & 0.9460 & 30.16 & 0.9007 & 29.32 & 0.9207 & 32.38 & 0.9378 & 31.54 & 0.9263 \\
      &   WeatherDiff\(_{64}\)~\cite{weather_diffusion}   & TPAMI'23 & 35.83 & 0.9566 & 30.09 & 0.9041 & 29.64 & 0.9312 & 30.71 & 0.9312 & 31.57 & 0.9308 \\
      & WeatherDiff\(_{128}\)~\cite{weather_diffusion}     & TPAMI`23 & 35.02 & 0.9516 & 29.58 & 0.8941 & 29.72 & 0.9216 & 29.66 & 0.9225 & 31.00 & 0.9225 \\
      &   AWRCP~\cite{AWRCP_YeCBSXJYCL23}  & ICCV'23  & 36.92 & 0.9652 & 31.92 & \textbf{0.9341} & 31.39 & 0.9329 & 31.93 & 0.9314 & 33.04 & 0.9409 \\
    & GridFormer~\cite{Gridformer}    & IJCV'24 & \uline{37.46} & 0.9640 & 31.71 & 0.9231 & 31.87 & 0.9335 & 32.39 & 0.9362 & 33.36 & 0.9392 \\
    &  MPerceiver~\cite{AiHZW024}  & CVPR'24  & 36.23 & 0.9571 & 31.02 & 0.9164 & 31.25 & 0.9246 & \textbf{33.21} & 0.9294 & 32.93 & 0.9319 \\
    &  DTPM~\cite{DTPM_0001CCXQL024}  & CVPR'24  & 37.01&   \uline{0.9663} & 30.92 & 0.9174 &  30.99  &  0.9340 & 32.72 & 0.9440 & 32.91 & 0.9404\\
    &  Histoformer~\cite{Histoformer_SunRGWC24} & ECCV'24 & 37.41 & 0.9656 & \uline{32.16} & 0.9261 & \uline{32.08} & \uline{0.9389} & \uline{33.06} & \uline{0.9441} & \uline{33.68} & \uline{0.9437} \\
    &  \cellcolor{my_color}\textbf{DSwinIR (Ours)}   & \cellcolor{my_color}2025 
    & \cellcolor{my_color}\textbf{38.11} & \cellcolor{my_color}\textbf{0.9683} 
    & \cellcolor{my_color}\textbf{32.58} & \cellcolor{my_color}\uline{0.9312} 
    & \cellcolor{my_color}\textbf{32.76} & \cellcolor{my_color}\textbf{0.9502} 
    & \cellcolor{my_color}{32.88} & \cellcolor{my_color}\textbf{0.9474} 
    & \cellcolor{my_color}\textbf{34.08} & \cellcolor{my_color}\textbf{0.9493} \\
     
    \bottomrule[1pt]
    \end{tabular}%
    }
\end{table*}
\begin{figure*}[!t] 
\centering \includegraphics[width=\linewidth]{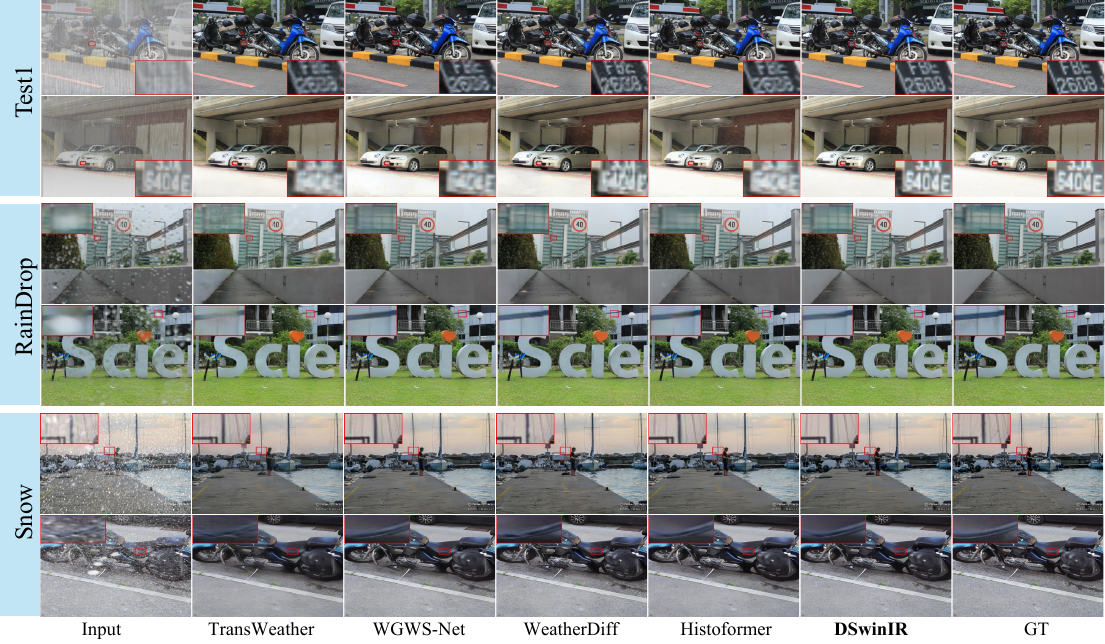}  
\caption{Visual results on the AllWeather datasets; from top to bottom are samples for outdoor-rain, raindrop and snow test data. Our DSwinIR achieves better clarity even when compared to the diffusion-based approaches.} 
\label{fig_allweather}  
\end{figure*}

\subsubsection{Compared Methods}
To comprehensively position DSwinIR within the literature, we conduct extensive comparisons against a spectrum of state-of-the-art models. This includes foundational, backbone-centric architectures such as Uformer \cite{Uformer}, Restormer \cite{Restormer}, DRSformer \cite{DRSformer}, GridFormer \cite{Gridformer}, {MambaIR~\cite{MambaIR}, VmambaIR~\cite{Shi2025VmambaIR}, and MaIR~\cite{Li2025MaIR}} allowing for a direct benchmark of the architectural efficacy of our proposed DSwin attention. Furthermore, we contrast our model with cutting-edge prompt-based and instruction-guided all-in-one methods, including AirNet \cite{AirNet}, PromptIR \cite{PromptIR}, InstructIR \cite{InstructIR}, AdaIR \cite{cui2025adair}, {PoolNet~\cite{PoolNet}, and VLUNet~\cite{Zeng2025VLUNet}}. Demonstrating superiority over this class of models highlights that our DSwinIR provides a more powerful and efficient solution than methods reliant on external semantic guidance or task-specific modules.

\subsubsection{Implementation Details}
DSwinIR is trained end-to-end using the AdamW optimizer with an initial learning rate of $2 \times 10^{-4}$, which is decayed using a cosine annealing schedule. Our model employs a four-level U-shaped architecture, with the number of DSwin Transformer blocks in the encoder stages set to [4, 6, 6, 8] from shallow to deep. We use a batch size of 8 and train on randomly cropped $128 \times 128$ patches from the respective training images. All experiments are conducted using PyTorch on 2 NVIDIA 3090 GPUs.

\subsection{Results and Analysis}
In this section, we present a detailed analysis of DSwinIR’s performance across a hierarchy of benchmarks, ranging from complex multi-degradation scenarios to specialized single-task settings. Our aim is not only to establish the state-of-the-art performance of our model but also to provide insights into the architectural and mechanistic factors that contribute to its effectiveness. In the following result tables, \textbf{Best} and \underline{second-best} results are highlighted.

\subsubsection{Results on All‑in‑One Image Restoration}
The most challenging test for any restoration model is the all-in-one setting, where a single network must handle a diverse array of degradations. We evaluate DSwinIR on the standard three-task and five-task benchmarks to demonstrate its capacity as a powerful universal backbone.

\paragraph{Performance on Three Degradation Tasks}
Following the classic benchmark setup \cite{AirNet,PromptIR}, we first evaluate on denoising, deraining, and dehazing. The quantitative results are presented in \Cref{tab_three_task}. DSwinIR achieves an average PSNR of 32.72 dB. This result demonstrates improvements over other foundational architectures like GridFormer (32.19 dB) and validates the efficacy of our DSwin attention. Furthermore, DSwinIR remains highly competitive even when compared to the latest prompt-based (e.g., AdaIR, 32.69 dB) and multimodal (e.g., VLUNet, 32.70 dB) methods. The ability of our vision-only backbone to perform at this level suggests its strong intrinsic representation power. The qualitative results in \Cref{fig_3_task} are consistent with these quantitative findings, showing that DSwinIR effectively restores fine textures and removes complex artifacts.

\paragraph{Performance on Five Degradation Tasks}
To further probe the model's generalization, we adopt the five-task benchmark, which introduces deblurring and low-light enhancement tasks. As detailed in \Cref{tab_five_task}, DSwinIR achieves an average PSNR of 30.20 dB. This result is highly competitive with the top-performing methods, such as the prompt-based AdaIR (30.20 dB). The model's performance is notably strong in tasks requiring significant spatial transformation, such as deraining (e.g., +0.93 dB over InstructIR) and dehazing (e.g., +2.99 dB over InstructIR). While its performance on low-light enhancement is competitive rather than leading, its overall strength across the other tasks secures this high average. This highlights the DSwinIR architecture's ability to jointly learn from a diverse set of degradations within a single set of weights.

\subsubsection{Results on Deweathering Datasets}
Beyond general restoration, we investigate DSwinIR's performance in the practical domain of deweathering, which requires handling complex, real-world atmospheric phenomena.

\paragraph{Synthetic Deweathering Benchmarks}
On the comprehensive AllWeather dataset collection, which includes snow, raindrops, and outdoor rain, DSwinIR demonstrates clear superiority. As presented in \Cref{tab_allweather}, our model achieves an average PSNR of {34.08 dB}, surpassing the previous best, Histoformer \cite{Histoformer_SunRGWC24}, by a convincing margin of 0.4 dB. The visual results in \Cref{fig_allweather} are consistent with the quantitative results. Compared to prior works, including diffusion-based models, DSwinIR consistently produces images with better clarity, more vibrant colors, and sharper details, effectively removing weather artifacts without introducing secondary distortions.

\begin{table}[!t]\small
\centering
\setlength\tabcolsep{7pt}
\renewcommand\arraystretch{1.25}

\caption{Comparison on de-weathering tasks on real-world datasets following \cite{weather_data}} \label{tab_real_weather}

\resizebox{0.4\textwidth}{!}{
\begin{tabular}{cccc}
\toprule[1pt]
Datasets & Methods & PSNR $\uparrow$ & SSIM $\uparrow$ \\
\hline
\multirow{4}{*}{ SPA+ } 
& Chen et al. \cite{ChenHTYDK22} &  37.32 & \uline{0.97} \\
& TransWeather \cite{Transweather}  & 33.64 & 0.93 \\
& WGWSNet \cite{weather_data} & \uline{38.94} & \textbf{0.98} \\
& \cellcolor{my_color}\textbf{DSwinIR (Ours)} & \cellcolor{my_color}\textbf{40.60} & \cellcolor{my_color}\textbf{0.98} \\
\hline 

\multirow{4}{*}{RealSnow} 
& Chen et al. \cite{ChenHTYDK22}&   29.37 & \uline{0.88} \\
& TransWeather  \cite{Transweather}    & 29.16 & 0.82 \\
& WGWSNet \cite{weather_data} & \uline{29.46} & 0.85 \\
&  \cellcolor{my_color}\textbf{DSwinIR (Ours)} & \cellcolor{my_color}\textbf{30.14} & \cellcolor{my_color}\textbf{0.89} \\
\hline 

\multirow{4}{*}{REVIDE} & Chen et al. \cite{ChenHTYDK22} &  20.10 & 0.85 \\
& TransWeather \cite{Transweather}  &  17.33 & 0.82 \\
& WGWSNet \cite{weather_data}  & \uline{20.44} & \underline{0.87} \\
& \cellcolor{my_color}\textbf{DSwinIR (Ours)}  &  \cellcolor{my_color}\textbf{20.80} & \cellcolor{my_color}\textbf{0.89} \\
\hline
\multirow{4}{*}{Average } 
& Chen et al. \cite{ChenHTYDK22} &  28.93 & \underline{0.90} \\
& TransWeather \cite{Transweather}  & 26.71  & 0.86 \\
& WGWSNet \cite{weather_data} & \underline{29.61} & \underline{0.90} \\
& \cellcolor{my_color}\textbf{DSwinIR (Ours)} & \cellcolor{my_color}\textbf{30.51} & \cellcolor{my_color}\textbf{0.92} \\
\bottomrule[1pt]

\end{tabular}
}

\end{table}

\begin{table*}[h!]
\centering
\footnotesize
\renewcommand{\arraystretch}{1.3}
\setlength{\tabcolsep}{3pt}
\caption{Quantitative comparison on the real-world WeatherBench dataset.}
\label{tab:weatherbench}
\resizebox{\linewidth}{!}{
{
\begin{tabular}{c|c|c|ccc|ccc|ccc|ccc}
\toprule[1pt]
{\multirow{2}{*}{\textbf{Type}}} & \multirow{2}{*}{\textbf{Method}} & \multirow{2}{*}{\textbf{Venue}} & \multicolumn{3}{c|}{\textbf{Dehaze}} & \multicolumn{3}{c|}{\textbf{Derain}} & \multicolumn{3}{c|}{\textbf{Desnow}} & \multicolumn{3}{c}{\textbf{Average}} \\ \cline{4-15} 
 & & & \textbf{PSNR}$\uparrow$ & \textbf{SSIM}$\uparrow$ & \textbf{LPIPS}$\downarrow$ & \textbf{PSNR}$\uparrow$ & \textbf{SSIM}$\uparrow$ & \textbf{LPIPS}$\downarrow$ & \textbf{PSNR}$\uparrow$ & \textbf{SSIM}$\uparrow$ & \textbf{LPIPS}$\downarrow$ & \textbf{PSNR}$\uparrow$ & \textbf{SSIM}$\uparrow$ & \textbf{LPIPS}$\downarrow$ \\ 
 \hline
{\multirow{7}{*}{\makecell{\rotatebox{90}{\textit{General}}}}} 
& DehazeFormer~\cite{DehazeFormer}  & TIP'23 & 24.12 & 0.745 & 0.345 & 36.05 & 0.954 & 0.181 & 28.88 & 0.849 & 0.178 & 29.68 & 0.849 & 0.235 \\
 & DCMPNet~\cite{Zhang2024DCMPNet}  & CVPR'24 & 21.18 & 0.506 & 0.491 & 32.04 & 0.876 & 0.282 & 24.81 & 0.614 & 0.546 & 26.01 & 0.665 & 0.440 \\  
 & DRSformer~\cite{DRSformer} & CVPR'23 & 19.95 & 0.694 & 0.404 & 33.98 & 0.943 & 0.209 & 28.00 & 0.836 & 0.197 & 27.31 & 0.824 & 0.270 \\
 & NeRD-Rain~\cite{Chen2024nerd-rain}  & CVPR'24 & 21.52 & 0.718 & 0.386 & 35.74 & 0.950 & 0.182 & 28.87 & 0.851 & 0.182 & 28.71 & 0.840 & 0.250 \\  
  & SnowFormer~\cite{Chen2022Snowformer}  & arXiv'22 & 22.71 & 0.736 & \underline{0.305} & 35.18 & 0.951 & 0.155 & 29.30 & 0.868 & \underline{0.143} & 29.06 & 0.852 & \underline{0.201} \\  
  & MPRNet~\cite{ZamirA2021Mprnet}  & CVPR'21 & 23.27 & 0.739 & 0.355 & 36.14 & 0.954 & 0.171 & 29.18 & 0.860 & 0.177 & 29.53 & 0.851 & 0.234 \\
   & Restormer~\cite{Restormer}  & CVPR'22 & 19.30 & 0.687 & 0.412 & 34.49 & 0.945 & 0.197 & 27.95 & 0.836 & 0.197 & 27.25 & 0.823 & 0.269 \\ \hline
\multirow{9}{*}{\makecell{\rotatebox{90}{\textit{All-in-One}}}} 
& AirNet~\cite{AirNet}  & CVPR'22 & 20.94 & 0.705 & 0.383 & 33.59 & 0.942 & 0.224 & 22.06 & 0.780 & 0.291 & 25.53 & 0.809 & 0.299 \\
 & TransWeather~\cite{Transweather} & CVPR'22 & 19.79 & 0.680 & 0.397 & 29.34 & 0.903 & 0.294 & 24.96 & 0.796 & 0.231 & 24.70 & 0.793 & 0.307 \\
  & WGWS-Net~\cite{weather_data} & CVPR'23 & 13.79 & 0.603 & 0.535 & \underline{37.08} & \textbf{0.961} & \textbf{0.117} & 20.81 & 0.780 & 0.248 & 23.89 & 0.781 & 0.300 \\
  & PromptIR~\cite{PromptIR} & NeurIPS'23 & 21.11 & 0.713 & 0.375 & 34.54 & 0.944 & 0.198 & 27.93 & 0.836 & 0.195 & 27.86 & 0.831 & 0.256 \\
  & DiffUIR~\cite{Zheng2024DiffUIR}& CVPR'24 & 22.74 & 0.744 & 0.355 & 35.93 & \underline{0.955} & 0.172 & 29.50 & {0.870} & {0.162} & 29.39 & \underline{0.856} & 0.230 \\
 & MWFormer~\cite{Zhu2024MWFormer}& TIP'24 & \underline{24.42} & \underline{0.746} & \textbf{0.284} & 35.15 & 0.951 & \underline{0.153} & \underline{29.98} & \underline{0.872} & \textbf{0.133} & \underline{29.85} & \underline{0.856} & \textbf{0.190} \\
 & Histoformer~\cite{Histoformer_SunRGWC24} & ECCV'24 & 17.69 & 0.669 & 0.437 & 30.70 & 0.916 & 0.279 & 25.39 & 0.808 & 0.225 & 24.59 & 0.798 & 0.314 \\
 & AdaIR~\cite{cui2025adair} & ICLR'25 & 23.08 & 0.731 & 0.351 & 34.87 & 0.946 & 0.192 & 28.44 & 0.837 & 0.179 & 28.80 & 0.838 & {0.240} \\
 \cline{2-15}
 
  & 
\cellcolor{my_color}\textbf{DSwinIR (Ours)} & \cellcolor{my_color}2025 & 
\cellcolor{my_color}\textbf{25.15} & \cellcolor{my_color}\textbf{0.766} & \cellcolor{my_color}0.329 & 
\cellcolor{my_color}\textbf{37.17} & \cellcolor{my_color}\textbf{0.961} & \cellcolor{my_color}0.173 & 
\cellcolor{my_color}\textbf{30.12} & \cellcolor{my_color}\textbf{0.873} & 
\cellcolor{my_color}0.170 & 
\cellcolor{my_color}\textbf{30.81} & \cellcolor{my_color}\textbf{0.867} & \cellcolor{my_color}0.224 \\
\bottomrule[1pt]
\end{tabular}
}
}
\end{table*}
\begin{figure*}[h!]
\centering
 \includegraphics[width=\linewidth]{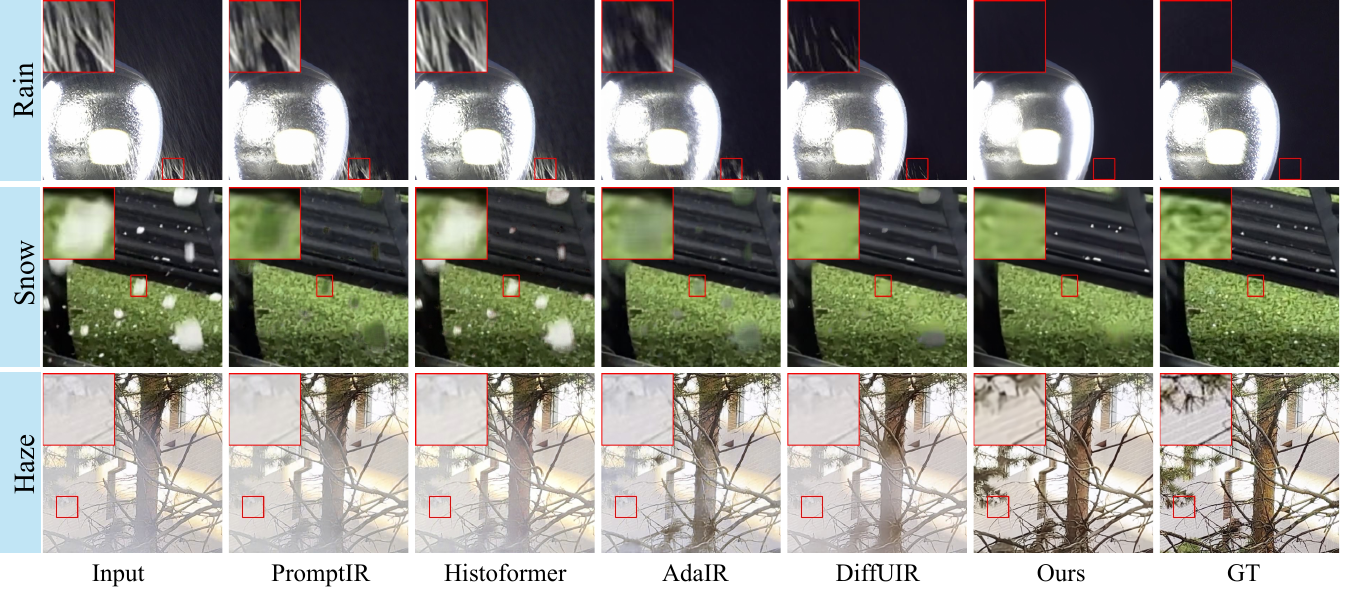}
\caption{{Visual results on the real-world WeatherBench dataset.} }
\label{fig:weatherbench}
\end{figure*}

\paragraph{Real-World Deweathering Benchmarks}
On a challenging benchmark composed of real-world rain, snow, and haze datasets \cite{weather_data}, DSwinIR again performs well. As shown in \cref{tab_real_weather}, our model achieves an average PSNR of {30.51 dB}, significantly outperforming the previous state-of-the-art WGWSNet \cite{weather_data} by 0.9 dB.

Moreover, we conduct evaluations on the recent, large-scale real-world WeatherBench dataset~\cite{Guan2025WeatherBench}, with results presented in \cref{tab:weatherbench}. Our method, DSwinIR, achieves competitive overall performance.
Notably, DSwinIR performs well in the Dehaze and Desnow tasks, outperforming the next-best method (MWFormer) by 0.73 dB in Dehaze PSNR and 0.14 dB in Desnow PSNR. While WGWS-Net shows strong performance in Derain, our model achieves a close second-best result. This performance across all categories results in a high overall average, indicating a good balance and robustness across diverse weather conditions. As illustrated in \cref{fig:weatherbench}, the qualitative results are consistent with these quantitative findings. DSwinIR effectively removes complex weather artifacts (e.g., dense fog, heavy snow) while preserving fine-grained details and avoiding color casts.

\begin{table*}[t]
\centering
\caption{Quantitative results on the Composite Degradation Setting.}
\label{tab_composite_cdd}
\renewcommand\arraystretch{1.2}
\resizebox{\linewidth}{!}{
\begin{tabular}{cccccccccccccc}
\toprule[1pt]
\textbf{Method} & \textbf{Venue}& \textbf{l} & \textbf{h} & \textbf{r}& \textbf{s}& \textbf{l+h} & \textbf{l+r} & \textbf{l+s}& \textbf{h+r} & \textbf{h+s} & \textbf{l+h+r} & \textbf{l+h+s} & \textbf{Avg.}\\
\midrule
Restormer~\cite{Restormer} & CVPR'22 & 26.29 & 28.35 & 33.10 & 33.43 & 24.80 & 25.28 & 24.99 & 26.80 & 26.15 & 23.90 & 23.82 & 26.99 \\
NAFNet~\cite{chen2022simple} & ECCV'22 & 24.50 & 25.34 & 29.24 & 29.54 & 21.91 & 22.75 & 22.79 & 23.67 & 23.86 & 21.03 & 20.82 & 24.13 \\
AirNet~\cite{AirNet} & CVPR'22 & 24.83 & 24.21 & 26.55 & 26.79 & 23.23 & 22.82 & 23.29 & 22.21 & 23.29 & 21.80 & 22.24 & 23.75 \\
PromptIR~\cite{PromptIR} & NeurIPS’23 & 26.32 & 26.10 & 31.56 & 31.53 & 24.49 & 25.05 & 24.51 & 24.54 & 23.70 & 23.74 & 23.33 & 25.90 \\
TransWeather~\cite{Transweather} & CVPR'22 & 23.39 & 23.95 & 26.69 & 25.74 & 22.24 & 22.62 & 21.80 & 23.10 & 22.34 & 21.55 & 21.01 & 23.13 \\
WeatherDiff~\cite{weather_data} & TPAMI'23 & 23.58 & 21.99 & 24.85 & 24.80 & 21.83 & 22.69 & 22.12 & 21.25 & 21.99 & 21.23 & 21.04 & 22.49 \\
WGWSNet~\cite{weather_data} & CVPR'23 & 24.39 & 27.90 & 33.15 & 34.43 & 24.27 & 25.06 & 24.60 & 27.23 & 27.65 & 23.90 & 23.97 & 26.96 \\
InstructIR~\cite{InstructIR} & ECCV'24 & \underline{26.70} & 32.61 & \underline{33.51} & 34.45 & 24.36 & 25.41 & \underline{25.63} & 28.80 & 29.64 & 24.84 & 24.32 & 28.21 \\
OneRestore~\cite{guo2024onerestore} & ECCV'24 & 26.55 & \underline{32.71} & 33.48 & \underline{34.50} & \underline{26.15} & \underline{25.83} & 25.56 & \underline{30.27} & \underline{30.46} & \underline{25.18} & \underline{25.28} & \underline{28.47} \\
\midrule
\cellcolor{my_color}\textbf{DSwinIR (Ours)} &
\cellcolor{my_color}2025 &
\cellcolor{my_color}{\textbf{27.38}} &
\cellcolor{my_color}{\textbf{33.72}} &
\cellcolor{my_color}{\textbf{34.95}} &
\cellcolor{my_color}{\textbf{35.44}} &
\cellcolor{my_color}{\textbf{26.56}} &
\cellcolor{my_color}{\textbf{26.54}} &
\cellcolor{my_color}{\textbf{26.24}} &
\cellcolor{my_color}{\textbf{30.94}} &
\cellcolor{my_color}{\textbf{30.54}} &
\cellcolor{my_color}{\textbf{25.70}} &
\cellcolor{my_color}{\textbf{25.70}} &
\cellcolor{my_color}{\textbf{29.43}}\\
\bottomrule[1pt]
\end{tabular}

}
\end{table*}
\begin{figure*}[!h]
\centering
 \includegraphics[width=\linewidth]{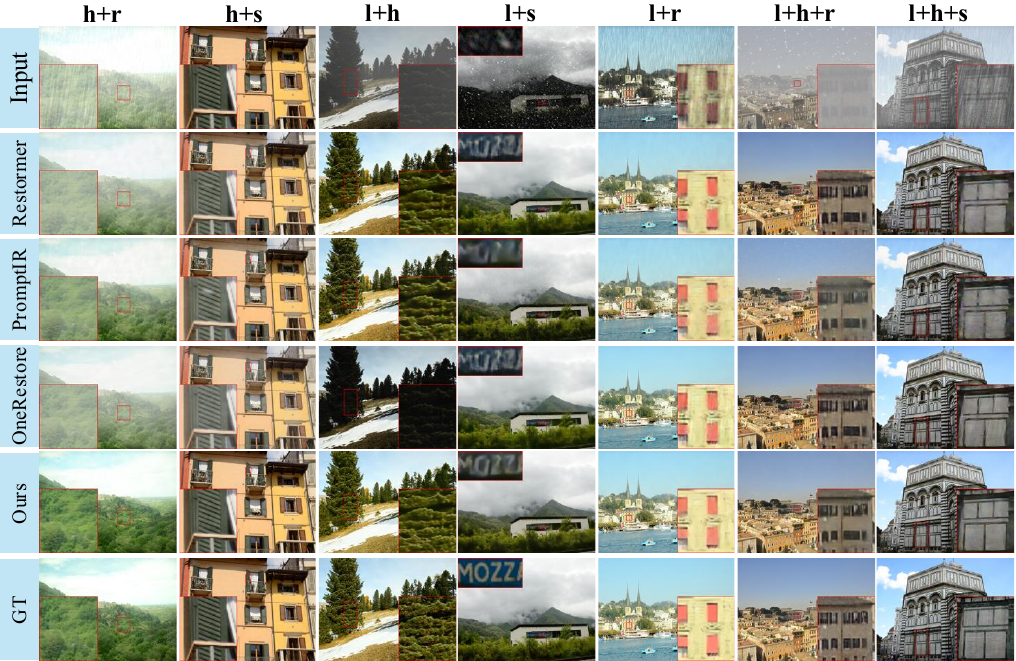}
\caption{Visual results on the composite degradation tasks, showcasing performance on mixed degradations. Our method more effectively mitigates multiple interacting degradations, restoring clearer images with improved color fidelity and detail.}
\label{fig:cdd}
\end{figure*}

\subsubsection{Results on Composite Degradations}
We evaluate DSwinIR on the challenging composite degradation benchmark (CDD-11) \cite{guo2024onerestore}, which features 11 combinations of low-light, haze, rain, and snow. As presented in \Cref{tab_composite_cdd}, our DSwinIR achieves an average PSNR of 29.43 dB, which is 0.96 dB higher than the next-best method, OneRestore \cite{guo2024onerestore}. The proposed DSwinIR can learn to adapt its receptive field to different degradation structures (e.g., thin rain streaks or broad haze), and our model achieves the advanced performance across all 11 individual and combined degradation settings, demonstrating the architecture's robustness in handling complex, interacting artifacts. The qualitative results in \Cref{fig:cdd} are consistent with these quantitative findings, showing that DSwinIR effectively removes interacting artifacts while preserving image detail.

\subsubsection{Results on Single‑Task Settings}
To complement the all-in-one results, we also investigate the performance of DSwinIR on single-task benchmarks. These experiments are designed to assess the model's effectiveness when its architecture is optimized for a single task, and to benchmark its performance against models developed specifically for these applications.

\paragraph{Deraining on Rain100L}
We first evaluate the model's performance on the standard Rain100L benchmark. As shown in \Cref{tab:single_deraining}, this is a highly competitive field. When trained solely for this task, DSwinIR demonstrates strong performance. The gain over other backbone architectures like PromptIR (+1.98 dB) suggests the architectural progress provided by our DSwin attention. Moreover, DSwinIR also achieves a 0.12 dB improvement over the recent AdaIR model.

\begin{table*}[h]
\centering
\caption{Single-task deraining results on the Rain100L dataset.}
\label{tab:single_deraining}
\renewcommand\arraystretch{1.15}
\resizebox{\linewidth}{!}{
\begin{tabular}{l|ccccccccccc}
\toprule
Method & DIDMDN \cite{zhang2018DIDMDN} & UMR \cite{yasarla2019UMR} & SIRR \cite{wei2019sirr} & MSPFN\cite{jiang2020multi} & LPNet\cite{LPNet} & AirNet\cite{AirNet} & Restormer\cite{Restormer} & PromptIR\cite{PromptIR} & AdaIR\cite{cui2025adair} & \cellcolor{my_color}\textbf{DSwinIR (Ours)} \\
\midrule
PSNR & 23.79 & 32.39 & 32.37 & 33.50 & 33.61 & 34.90 & 36.74 & 37.04 & \uline{38.90} & \cellcolor{my_color}\textbf{39.02} \\ 
SSIM & 0.773 & 0.921 & 0.926 & 0.948 & 0.958 & 0.977 & 0.978 & 0.979 & \uline{0.985} & \cellcolor{my_color}\textbf{0.986} \\
\bottomrule
\end{tabular}
}
\end{table*}

\paragraph{Dehazing on SOTS Outdoor}
The task of single-image dehazing presents a different challenge, often requiring a nuanced understanding of both global atmospheric light and local object details. As detailed in \Cref{tab:single_dehazing}, DSwinIR achieves a PSNR of 32.89 dB. This result represents a 1.09 dB gain over the next-best method, AdaIR. We posit that the deformable mechanism's ability to adapt its receptive field is well-suited to the non-homogeneous nature of haze, while the sliding mechanism's capacity to preserve fine structural details ensures object clarity.
\begin{table*}[h]
\centering
\caption{Single-task dehazing results on the SOTS outdoor test set.}
\label{tab:single_dehazing}
\resizebox{0.95\linewidth}{!}{%
\begin{tabular}{l|cccccccccc}
\toprule
Method & DehazeNet \cite{DehazeNet} & MSCNN \cite{ren2016mscnn} & AODNet \cite{AODNet} & EPDN \cite{qu2019epdn} & FDGAN \cite{FDGAN} & AirNet \cite{AirNet} & Restormer \cite{Restormer} & PromptIR \cite{PromptIR} & AdaIR \cite{cui2025adair} & \cellcolor{my_color}\textbf{DSwinIR (Ours)} \\
\midrule
PSNR & 22.46 & 22.06 & 20.29 & 22.57 & 23.15 & 23.18 & 30.87 & 31.31 & \uline{31.80} & \cellcolor{my_color}\textbf{32.89} \\
SSIM & 0.851 & 0.908 & 0.877 & 0.863 & 0.921 & 0.900 & 0.969 & 0.973 & \uline{0.981} & \cellcolor{my_color}\textbf{0.984} \\
\bottomrule
\end{tabular}
}
\end{table*}

\paragraph{Real-World Deraining on SPA-Data}
We also test DSwinIR on the real-world SPA-Data benchmark. The results, summarized in \Cref{tab:single_deraining_spa}, show DSwinIR achieves a PSNR of 49.19 dB. This represents a 0.66 dB gain over the strong DRSformer model. It suggests that the DSwin attention's ability to learn adaptive, data-driven receptive fields provides a practical advantage for tackling such complex, non-uniform degradations.

\begin{table}[h]
\centering
\caption{ Quantitative comparison on the LOLv2 dataset \cite{Yang2021LOLv2} for low-light enhancement.}
\label{tab:lolv2}
\setlength{\tabcolsep}{6pt}
{
\begin{tabular}{l|cc|cc}
\toprule
\multirow{2}{*}{Method} & \multicolumn{2}{c|}{LOLv2-Real} & \multicolumn{2}{c}{LOLv2-Syn} \\
& PSNR  & SSIM $\uparrow$ & PSNR $\uparrow$ & SSIM   \\
\midrule
RetinexNet~\cite{Wei2018RetinexNet} & 15.47 & 0.567 & 17.13 & 0.798 \\
UFormer~\cite{Uformer} & 18.82 & 0.771 & 19.66 & 0.871 \\
FIDE~\cite{Xu2020FIDE} & 16.85 & 0.678 & 15.20 & 0.612 \\
DRBN~\cite{Yang2020DRBN} & 20.29 & 0.831 & 23.22 & 0.927 \\
KinD~\cite{Zhang2021KinD} & 14.74 & 0.641 & 13.29 & 0.578 \\
Restormer~\cite{Restormer} & 19.94 & 0.827 & 21.41 & 0.830 \\
MIRNet~\cite{Zamir2020MIRNet} & 20.02 & 0.820 & 21.94 & 0.876 \\
SNR-Net~\cite{Xu2022SNRNET} & 21.48 & \underline{0.849} & 24.14 & 0.928 \\
FourLLIE~\cite{Wang2023FourLLIE} & 21.60 & 0.847 & 24.17 & 0.917 \\
Retinexformer~\cite{Retinexformer} & \underline{22.80} & 0.840 & \underline{25.67} & \underline{0.930} \\
\midrule
\rowcolor{my_color} \textbf{DSwinIR (Ours)} & \textbf{23.17} & \textbf{0.851} & \textbf{25.74} & \textbf{0.934} \\
\bottomrule
\end{tabular}
}
\end{table}

\begin{table*}[h]
\centering
\caption{Quantitative comparison for the mixed task (low-light enhancement and deblurring) on the LowBlur dataset \cite{zhou2022lowblur}.}
\label{tab:lowblur}
\renewcommand\arraystretch{1.1}  
\setlength{\tabcolsep}{4pt}  
\resizebox{\linewidth}{!}{ 
{
\begin{tabular}{l|cccccccccc}
\toprule
Method & KinD++~\cite{Zhang2021KinD} & DRBN~\cite{Yang2020DRBN} & DeblurGAN-v2~\cite{Kupyn2019Deblurgan} 
& MIMO~\cite{Cho2021MIMO} & NAFNet~\cite{chen2022simple} & LEDNet~\cite{Zhou2022LEDNet} & RetinexFormer~\cite{Retinexformer} & Restormer~\cite{Restormer} & DarkIR~\cite{Feijoo2025DarkIR} & \cellcolor{my_color}\textbf{DSwinIR (Ours)} \\
\midrule
PSNR   & 21.26 & 21.78 & 22.30 & 22.41 & 25.36 & 25.74 & 26.02 & 26.72 & \underline{27.30} & \cellcolor{my_color}\textbf{27.33} \\
SSIM  & 0.753 & 0.768 & 0.745 & 0.835 & 0.882 & 0.850 & 0.887 & \underline{0.902} & 0.898 & \cellcolor{my_color}\textbf{0.909} \\
\bottomrule
\end{tabular}
}
} 
\end{table*}

{
\paragraph{Low-Light Image Enhancement}
We first evaluate DSwinIR on the widely-used low-light enhancement benchmark, LOLv2, which includes both real-world (LOLv2-Real) and synthetic (LOLv2-Syn) test sets. As presented in Table~\ref{tab:lolv2}, our model achieves state-of-the-art performance on both subsets.  
Furthermore, to assess the model's capability in handling complex, compound degradations, we evaluate DSwinIR on the mixed task of joint low-light enhancement and deblurring. Following the experimental setup of DarkIR~\cite{Feijoo2025DarkIR} on the LowBlur dataset, our model again demonstrates superior performance, as detailed in Table~\ref{tab:lowblur}. DSwinIR achieves the highest scores with 27.33 dB in PSNR and 0.909 in SSIM. These results underscore the strong generalization capability of our model for the dedicated low-light enhancement task.}

\begin{table*}[h]
\centering
\caption{Single-task deraining results on the real-world SPA-Data dataset.}
\label{tab:single_deraining_spa}
\renewcommand\arraystretch{1.15}
\resizebox{\linewidth}{!}{
\begin{tabular}{l|ccccccccc}
\toprule
Method & MSPFN \cite{jiang2020multi} & MPRNet \cite{ZamirA2021Mprnet} & DualGCN \cite{li2021dual} & SPDNet \cite{huang2017spdnet} & Uformer \cite{Uformer} & Restormer \cite{Restormer} & IDT \cite{Xiao2023IDT}& DRSformer \cite{DRSformer} & \cellcolor{my_color}\textbf{DSwinIR (Ours)} \\
\midrule
PSNR & 43.43 & 43.64 & 44.18 & 43.20 & 46.13 & 47.98 & 47.35 & \uline{48.53} & \cellcolor{my_color}\textbf{49.19} \\
SSIM & 0.9843 & 0.9844 & 0.9902 & 0.9871 & 0.9913 & 0.9921 & \uline{0.9930} & 0.9924 & \cellcolor{my_color}\textbf{0.9938} \\
\bottomrule
\end{tabular}
}
\end{table*}

\begin{table}[h!]
\caption{Efficiency evaluation of different methods with 256$\times$256 input.}
\label{tab:efficience}
\centering
\resizebox{0.985\linewidth}{!}{%
\begin{tabular}{@{}lccccc@{}}
\toprule
Methods        & Core Operation                    & Params. & FLOPs & Latency & PSNR \\ \midrule
    AirNet \cite{AirNet}        & Deformable Conv                 & 9M     & 301G  & 0.57s    & 31.20 \\
    Uformer \cite{Uformer}        & Window Attention                & 20M     & 41G  & 0.45s    & 30.66 \\
    Restormer \cite{Restormer}  & Transposed Self-Attention        & 26M     & 155G  & 0.67s    & 30.75 \\
    PromptIR \cite{PromptIR}   & Trans. SA + Prompts             & 36M     & 173G  & 0.71s    & 32.06 \\
    DRSformer \cite{DRSformer}    & Sparse Window Attention        & 33M     & 243G  & 0.82s    & 31.26 \\
\rowcolor{my_color} 
\textbf{DSwinIR (Ours)} & \textbf{DSwin Attention} & \textbf{24M} & \textbf{147G} & \textbf{0.55s} & \textbf{32.72} \\ \bottomrule
\end{tabular}%
}
\end{table}

\subsection{Ablation Study}
\label{sec:ablation}
\begin{figure}[!t] 
\centering 
\includegraphics[width=0.85\linewidth]{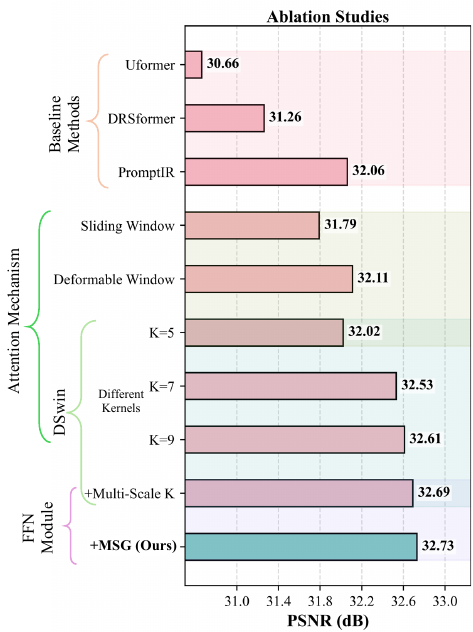} 
\caption{Ablation studies demonstrating the effectiveness of our key components. We evaluate (1) different attention mechanisms, showing improvements from sliding window (31.79 dB) and Deformable Window (32.11 dB) over baselines; (2) DSwin configurations with various kernel sizes ($K$=5, 7, 9) and multi-scale enhancement (32.69 dB); and (3) FFN module with MSG enhancement, achieving the best performance (32.72 dB). All experiments report the average PSNR of three distinct degradation tasks.} 
\label{fig_ablation_plot} 
\end{figure}

To validate our design choices and rigorously quantify the contribution of each proposed component, we conduct a series of in-depth ablation studies. We use the challenging three-task all-in-one setting as our testbed and present the comprehensive results in \Cref{fig_ablation_plot}.

\paragraph{Efficacy of the Core DSwin Mechanisms}
To validate our design, we incrementally introduce our components starting from established baselines. The Uformer \cite{Uformer} and DRSformer \cite{DRSformer} models provide benchmarks at 30.66 dB and 31.26 dB, respectively. Our first experiment, isolating the {sliding window} mechanism, yields 31.79 dB, confirming the effectiveness of the token-centric paradigm. In parallel, isolating the {deformable window} mechanism achieves 32.11 dB, a result that already outperforms the prompt-based PromptIR baseline (32.06 dB). When these two components are combined, the resulting {DSwin} module ($K$=7) achieves 32.53 dB. This performance gain is greater than the sum of the individual component improvements, demonstrating a clear synergistic effect.

\begin{figure}[h]
    \centering
    \includegraphics[width=\linewidth]{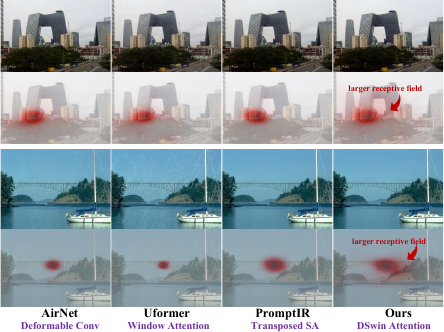}
    \caption{Visualization of receptive fields of multiple backbones by LAM\cite{lam}.}
\label{fig:erf_and_efficiency}
\end{figure}

\paragraph{Effective Receptive Field and Efficiency}
The visualizations of the receptive field in \cref{fig:erf_and_efficiency} offer profound qualitative insights into why DSwinIR is so effective. Its receptive field is demonstrably both {global in reach and local in precision}. This is the direct result of our design principles: the sliding mechanism ensures a connected, expansive base, while the deformable mechanism learns to dynamically concentrate attention on areas like object contours and textures. 

Moreover, DSwinIR also exhibits computational efficiency, as detailed in \cref{tab:efficience}. It is crucial to note that DSwinIR achieves its superior performance without substantially increasing computational cost. Compared to its top-performing competitors, DSwinIR is significantly more economical. For instance, it requires 15\% fewer FLOPs and is 22\% faster than PromptIR, while delivering a substantially higher PSNR. The advantage is even more dramatic when compared to DRSformer, against which DSwinIR uses about 40\% fewer FLOPs.

\paragraph{Impact of Kernel Size and Multi-Scale Strategy}
Having validated the DSwin concept, we next investigate its optimal configuration. We first analyze the impact of a fixed kernel size, testing $K$=5, $K$=7, and $K$=9. The results show a clear trend: performance improves from 32.02 dB ($K$=5) to 32.53 dB ($K$=7), and then begins to plateau at 32.61 dB ($K$=9). This suggests a trade-off, where a larger receptive field is beneficial, but excessively large kernels may not provide proportionally greater returns for their computational cost. This observation motivates our multi-scale strategy. By employing {multi-scale kernels} within a single attention module (+Multi-Scale K), we reach a PSNR of 32.69 dB. This result is superior to even the best-performing single-kernel model ($K$=9), providing clear evidence that empowering the model to simultaneously perceive features at multiple scales is more effective than relying on a single, large-but-fixed receptive field.

\paragraph{Contribution of the Multi-Scale Gated FFN}
Finally, we analyze the feed-forward network. Incorporating our proposed {multi-scale gated FFN} (+MSG) yields the model's peak performance at 32.72 dB. While this incremental gain is modest, it confirms the benefit of maintaining a holistic multi-scale philosophy across the entire Transformer block.

\subsection{Discussion and Limitation}

The extensive experiment results support the view that redesigning window attention with sliding and deformable sampling is effective across various benchmarks. Despite its demonstrated strengths, our work also illuminates several promising avenues for future research. First, while our architecture excels at restoring spatial structure, its performance on tasks dominated by extreme photometric adjustments, such as low-light enhancement, is competitive rather than leading. A promising future direction is to integrate explicit global photometric modeling, for instance, by combining our powerful backbone with lightweight techniques like 3D Look-Up Tables (3D-LUTs) or by incorporating a learnable codebook prior.
Second, our focus on architectural innovation means adapting DSwinIR to new tasks requires full retraining. A compelling direction is to integrate our backbone with parameter-efficient fine-tuning (PEFT) techniques, creating systems that are both powerful and highly adaptable. Finally, we believe the core principles of token-centric, content-adaptive attention hold significant potential beyond restoration, offering a promising new building block for other dense prediction tasks like semantic segmentation and object detection, where the ability to dynamically model features is paramount.

\section{Conclusion}\label{sec:conclusion}
In this paper, we revisit window-based self-attention for image restoration and propose the Deformable Sliding Window (DSwin) attention and the DSwinIR backbone. The token-centric sliding window reduces boundary effects by centering each token within its local context, while deformable sampling adapts the receptive field to image content. Extensive evaluations on all-in-one and single-task benchmarks show consistent improvements, including state-of-the-art results on several datasets. We hope DSwin will serve as a useful building block for future image restoration architectures and related dense prediction tasks.

\small
\bibliographystyle{IEEEtran}
\bibliography{main}

\begin{IEEEbiography}
[{\includegraphics[width=1.0in,height=1.25in,clip,keepaspectratio]{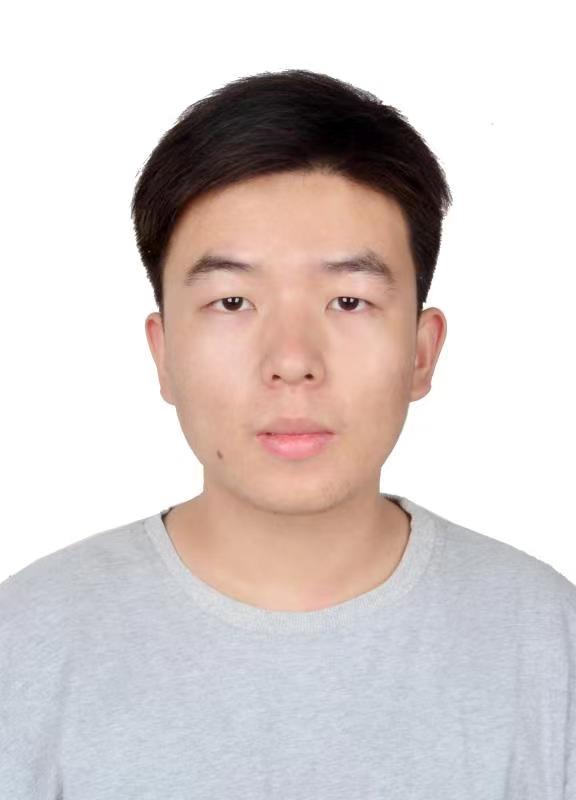}}]{Gang Wu} received the B.E. degree in the School of Computer Science and Technology from Soochow University, Jiangsu, China, in 2020. He is currently pursuing the Ph.D. degree in Faculty of Computing at Harbin Institute of Technology. His research interests include image restoration, representation learning, and self-supervised learning.
\end{IEEEbiography}

\begin{IEEEbiography}[{\includegraphics[width=1.0in,height=1.25in,clip,keepaspectratio]{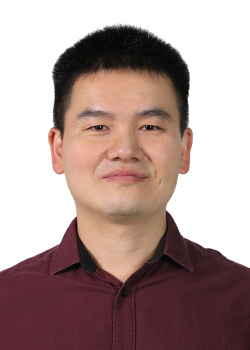}}]{Junjun Jiang}
received the B.S. degree in Mathematics from the Huaqiao University, Quanzhou, China, in 2009, and the Ph.D. degree in Computer Science from the Wuhan University, Wuhan, China, in 2014. 

From 2015 to 2018, he was an Associate Professor with the School of Computer Science, China University of Geosciences, Wuhan. From 2016 to 2018, he was a Project Researcher with the National Institute of Informatics (NII), Tokyo, Japan. He is currently a Professor with the School of Computer Science and Technology, Harbin Institute of Technology, Harbin, China. His research interests include image processing and computer vision. 
\end{IEEEbiography}

\begin{IEEEbiography}[{\includegraphics[width=1.0in,height=1.25in,clip,keepaspectratio]{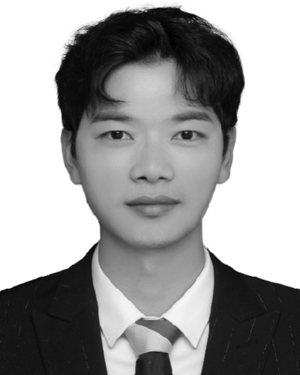}}]{Kui Jiang}
received the M.E. and Ph.D. degrees from the School of Computer Science, Wuhan University, Wuhan, China, in 2019 and 2022, respectively. Before July 2023, he was a Research Scientist with the Cloud BU, Huawei. He is currently an Associate Professor with the School of Computer Science and Technology, Harbin Institute of Technology. He received the 2022 ACM Wuhan Doctoral Dissertation Award. His research interests include image/video processing and computer vision.
\end{IEEEbiography}

\begin{IEEEbiography}[{\includegraphics[width=1.0in,height=1.25in,clip,keepaspectratio]{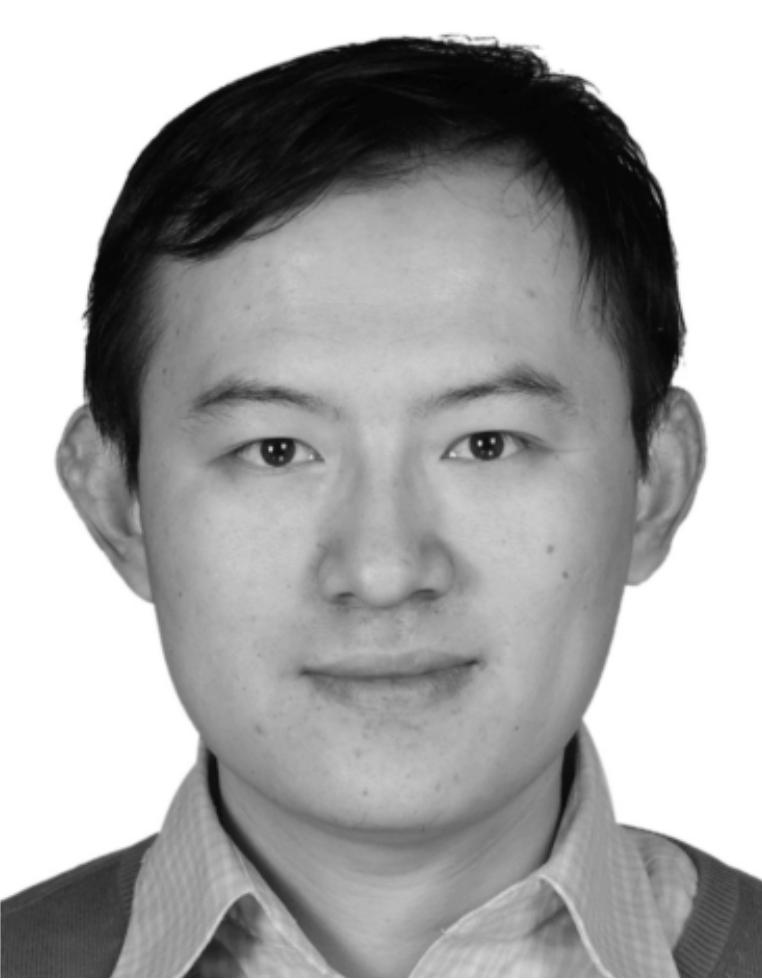}}]{Xianming Liu}
received the B.S., M.S., and Ph.D. degrees in computer science from the Harbin
Institute of Technology (HIT), Harbin, China,
in 2006, 2008, and 2012, respectively. In 2011, he spent half a year at the Department of Electrical and Computer Engineering, McMaster University, Canada, as a Visiting Student, where he was a Post-Doctoral Fellow from 2012 to 2013. He was a Project Researcher with the National Institute of Informatics (NII), Tokyo, Japan, from 2014 to 2017.
He is currently a Professor with the School of Computer Science and Technology, HIT. His research interests include trustworthy AI, 3D signal processing and biomedical signal processing.
\end{IEEEbiography}

\begin{IEEEbiography}[{\includegraphics[width=1.0in,clip,keepaspectratio]{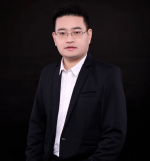}}]{Liqiang Nie}
received the B.Eng. and Ph.D. degree from Xi’an Jiaotong University and National University of Singapore (NUS), respectively. After Ph.D., he continued his research in NUS as a research fellow for three years. He is currently the dean of the Department of Computer Science and Technology, Harbin Institute of Technology (Shenzhen). His research interests lie primarily in multimedia computing and information retrieval. He is an AE of IEEE TIP, IEEE TKDE, IEEE TMM, IEEE TCSVT, ACM ToMM, and Information Science. Meanwhile, he serves as the chair of ICMR 2025, ICME 2025, and ACM MM 2027. He is a member of ICME steering committee. He has received many awards over the past four years, like SIGMM rising star in 2020, MIT TR35 China 2020, SIGIR best student paper in 2021, IEEE AI’s 10 to Watch in 2022, ACM MM Best paper award in 2022, and the National Youth Science and Technology Award in 2024.
\end{IEEEbiography}

\end{document}